\documentclass[11pt,a4paper]{article}
\usepackage{amsfonts}
\usepackage{bbm}
\usepackage{tcolorbox}
\usepackage{subcaption}
\usepackage[final]{acl}
\usepackage[table]{xcolor}
\usepackage{times}
\usepackage{fontawesome5}
\usepackage{hyperref}
\usepackage{latexsym}
\usepackage{amsmath}
\usepackage{booktabs}
\usepackage{multirow}
\usepackage[T1]{fontenc}

\usepackage[utf8]{inputenc}

\usepackage{microtype}

\usepackage{inconsolata}

\usepackage{graphicx}

\title{VL-Calibration: Decoupled Confidence Calibration for
\\ Large Vision-Language Models Reasoning}

\author{
 \textbf{Wenyi Xiao\thanks{Equal Contribution}\textsuperscript{1}},
 ~\textbf{Xinchi Xu\footnotemark[1]\textsuperscript{1}},
 ~\textbf{Leilei Gan\thanks{Corresponding Author}\textsuperscript{1}}
\\
 \textsuperscript{1}Zhejiang University
\\
 \small{
   \{\href{mailto:wenyixiao@zju.edu.cn}{wenyixiao},
   \href{mailto:leileigan@zju.edu.cn}{leileigan}\}@zju.edu.cn
 }
}

\begin{document}
\maketitle

\begin{abstract}
Large Vision Language Models (LVLMs) achieve strong multimodal reasoning but frequently exhibit hallucinations and incorrect responses with high certainty, which hinders their usage in high-stakes domains. Existing verbalized confidence calibration methods, largely developed for text-only LLMs, typically optimize a single holistic confidence score using binary answer-level correctness. This design is mismatched to LVLMs: an incorrect prediction may arise from perceptual failures or from reasoning errors given correct perception, and a single confidence conflates these sources while visual uncertainty is often dominated by language priors. To address these issues, we propose \textbf{VL-Calibration}\footnote{\faGithub\ \href{https://github.com/Mr-Loevan/VL-Calibration}{github.com/Mr-Loevan/VL-Calibration}}, a reinforcement learning framework that explicitly decouples confidence into visual and reasoning confidence. To supervise visual confidence without ground-truth perception labels, we introduce an intrinsic visual certainty estimation that combines (i) visual grounding measured by KL-divergence under image perturbations and (ii) internal certainty measured by token entropy. We further propose token-level advantage reweighting to focus optimization on tokens based on visual certainty, suppressing ungrounded hallucinations while preserving valid perception.
Experiments on thirteen benchmarks show that VL-Calibration effectively improves calibration while boosting visual reasoning accuracy, and it generalizes to out-of-distribution benchmarks across model scales and architectures.

\end{abstract}

\begin{figure}[!h]
    \centering
    \includegraphics[width=1.0\linewidth]{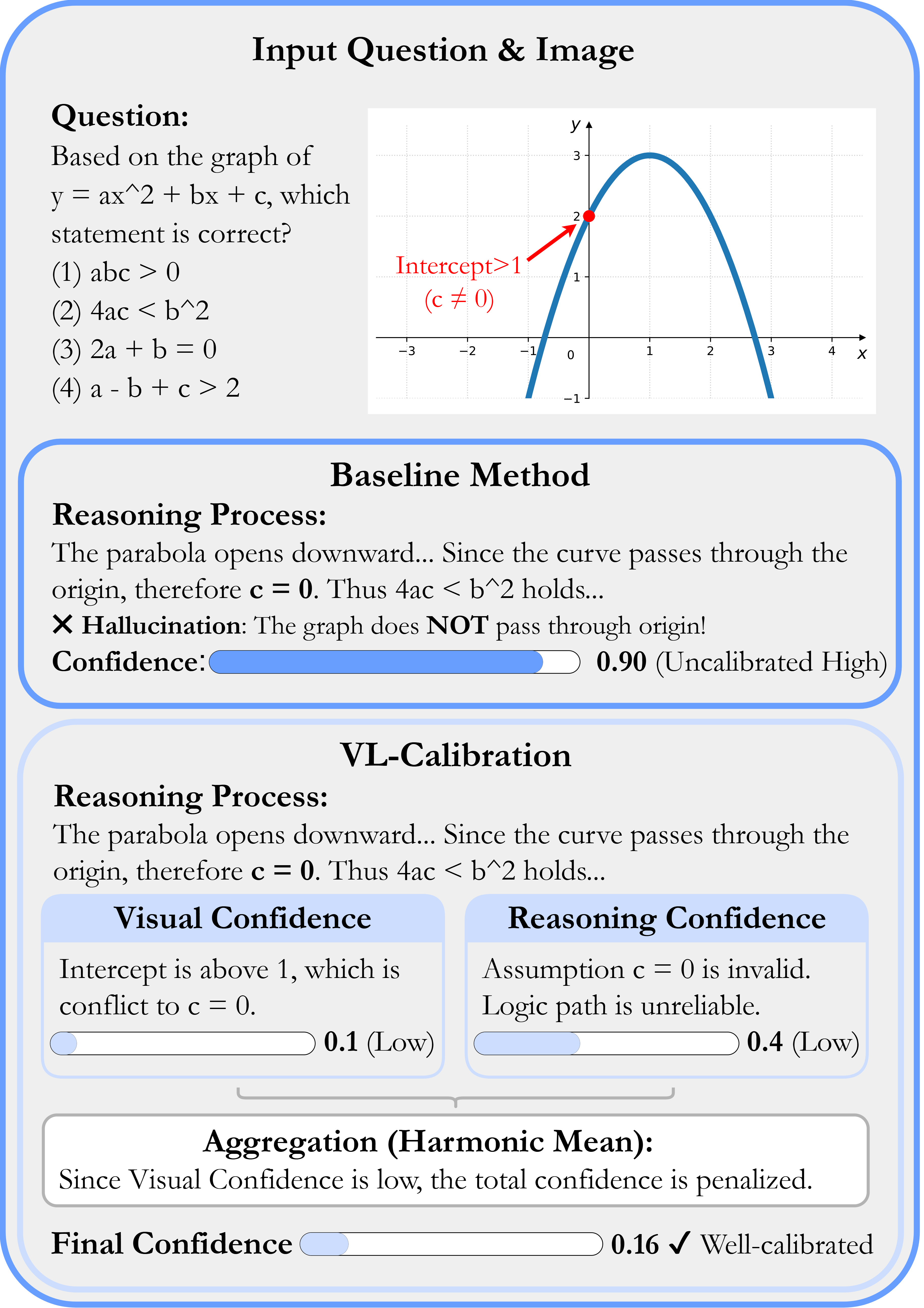}
    \caption{The baseline method (upper) makes overconfident assumptions. Instead, our method (lower) decouples confidence into visual and reasoning confidence, with clear identification of uncertainty sources and improved calibration.}
    \label{fig:uncertainty_badcase}
\end{figure}

\section{Introduction}
\label{intro}

Large Vision-Language Models (LVLMs) have demonstrated impressive capabilities in bridging visual perception and logical reasoning \citep{Qwen3-VL, wang2025internvl3_5,zhang2026fuse}. Despite their success, these models often exhibit severe hallucination that typically generate factually incorrect responses~\citep{ptrue,verbalize}, limiting their usage in high-stakes domains such as healthcare or law~\citep{xiao2025detecting,hu2025fine,shi2025revealer}.

\begin{figure*}[t]
    \centering
    \includegraphics[width=1.0\textwidth]{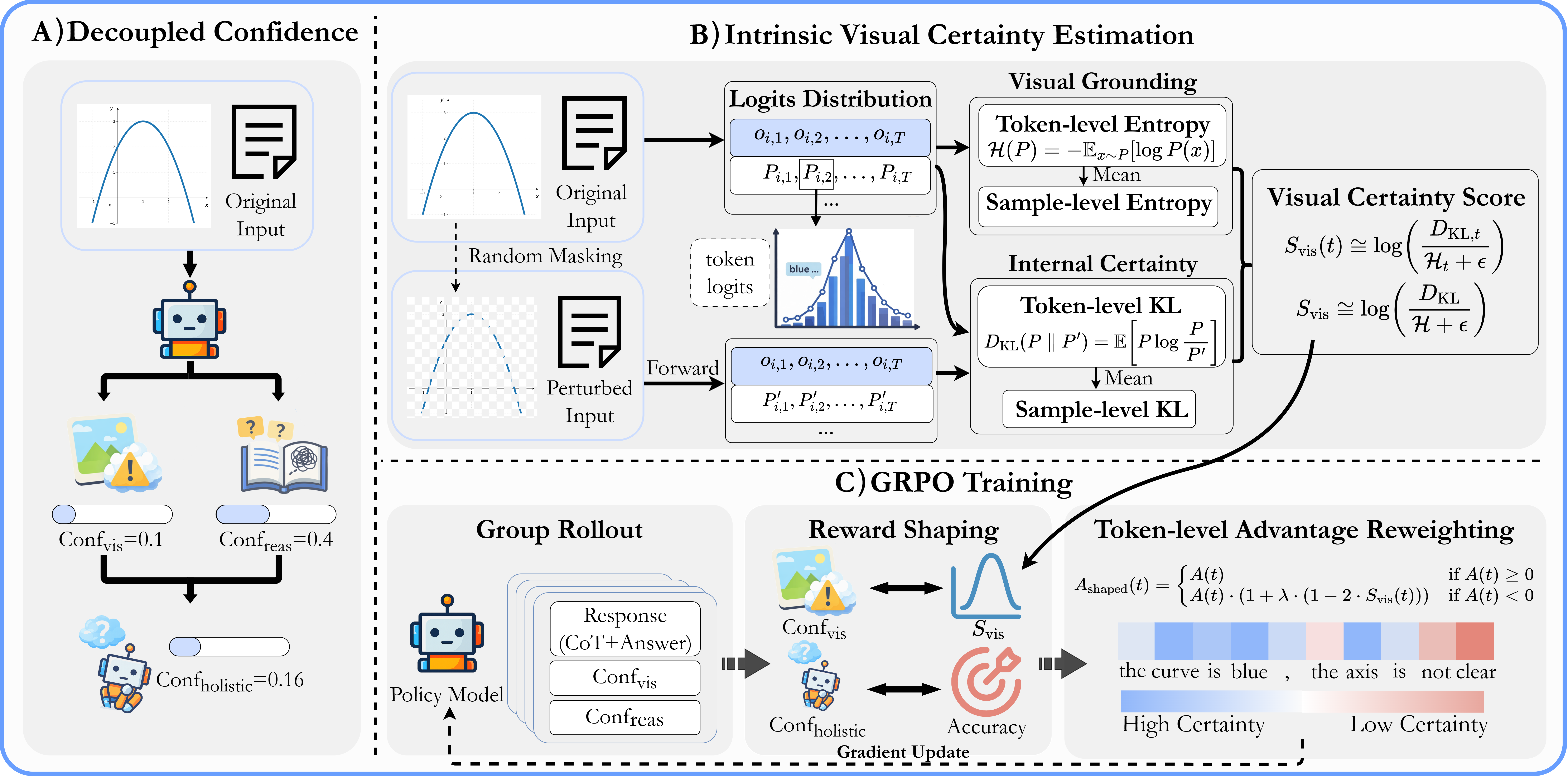}
    \caption{\textbf{Overview of our framework.}
    \textbf{A) Decoupled Confidence Inference.} The LVLM explicitly outputs separate visual and reasoning confidence to derive a holistic confidence.
    \textbf{B) Intrinsic Visual Certainty Estimation.} We quantify visual certainty by measuring visual grounding and internal certainty.
    \textbf{C) GRPO Training.} We align visual confidence with visual certainty score , and the holistic confidence with the answer accuracy. Additionally, we apply token-level advantage reweighting based on token visual certainty.
    }
    \label{fig:pipeline}
\end{figure*}

One solution to overcome the aforementioned challenge is to teach models to verbalize their confidence~\citep{ptrue,verbalize} (\textit{e.g.}, "my confidence is 8/10") alongside the answers.
Recent work has explored \textbf{verbalized confidence calibration} in large language models (LLMs)\footnote{For brevity, we use verbalized confidence calibration and calibration in this paper alternatively.}~\citep{rewardingdoubt,ppo-c}.
For example, SaySelf~\citep{sayself} trains models to output a verbalized confidence alongside an answer via distilled rationale from GPT-4 and then uses reinforcement learning to align confidence with accuracy.
While effective, the performance is limited by the capability of GPT-4.
To solve this, RLCR~\citep{RLCR} trains models to output verbalized confidence via RL using the Brier Score~\citep{brier}, i.e., L2 distance of predicted confidence and binary accuracy, and simultaneously incentivizes reasoning capability through accuracy reward.

However, directly extending verbalized confidence calibration to LVLM reasoning faces several fundamental challenges.
First, in LVLMs, an incorrect answer may arise from \emph{perceptual hallucinations} (misreading or ignoring the image) or from \emph{reasoning errors} given a correct perception.
Consequently, a single confidence score conflates these error sources in LVLMs, thereby hindering precise error localization.
Second, recent studies indicate LVLM reasoning is often dominated by language priors~\citep{ramakrishnan2018overcoming, jing2020overcoming,mathverse}.
Consequently, the intrinsic visual uncertainty is likely overshadowed by these language priors, leading to incorrect overall confidence calibration.

To address these limitations, we introduce \textbf{VL-Calibration}, a verbalized confidence calibration framework that decouple a single confidence into regarding visual perception from that of logical reasoning.
As shown in Figure~\ref{fig:pipeline} (a), we structure calibration into two phases and elicit separate confidence tokens for the visual rationale and the reasoning chain.
By doing so, VL-Calibration enables confidence expression for both perception and reasoning, and also identify clearly the uncertainty source.

We employ reinforcement learning to train VL-Calibration.
However, during training, this decoupling framework faces the lack of ground-truth labels for visual perception confidence.
Existing uncertainty estimation methods for LLMs and LVLMs largely fall into two categories.
\textbf{Sampling-based methods}, such as Self-Consistency~\citep{self-consistency} and VL-Uncertainty~\citep{zhang2024vl-uncertainty}, infer uncertainty by aggregating multiple generations, but incur substantial computational overhead.
\textbf{Internal-state methods} leverage logits \citep{ptrue} or hidden representations \citep{CoCoA} to predict correctness.
For example, Self-Certainty~\citep{Self-Certainty} measures KL-divergence of logits distribution from uniform distribution.
However, they overlook the visual grounding characteristic of LVLMs.

To address the absence of ground-truth labels for visual confidence, we propose estimating the certainty of visual perception by simultaneously considering \textbf{Visual Grounding} and \textbf{Internal Certainty}.
Specifically, to measure visual grounding, we compute the KL-divergence between the model's output distributions given the original image versus a perturbed image.
A higher KL-divergence indicates that the model is sensitive to visual content, implying strong grounding.
Second, to quantify internal uncertainty, we calculate the token entropy of the visual description, where lower entropy reflects higher model confidence.
We integrate the two estimations via a log-scale formulation to derive the \textbf{Visual Certainty Reward}, which enjoys the merits of (i) rewarding responses that are both visually responsive and internally confident; and (ii) optimization stability by compressing the dynamic numeric range to facilitate stable RL training.
Finally, to overcome the weakness that binary outcome reward treats each token equally, we propose \textbf{Token-level Advantage Reweighting}, which leverages the aforementioned visual certainty estimation to reweight advantage on high visual uncertainty tokens with negative advantage, thereby discouraging ungrounded hallucinations while preserving valid visual perception.

Extensive experiments across thirteen benchmarks demonstrate the effectiveness of our approach. VL-Calibration reduces the Expected Calibration Error (ECE) on Qwen3-VL models from 0.421 to 0.098 and simultaneously improves average accuracy by 2.3\%--3.0\% over the strongest baselines. We further observe consistent gains across model scales,  model architectures, and out-of-distribution benchmarks, validating the effectiveness of our proposed paradigm.

\section{Related Work}

\paragraph{LLM and LVLM Uncertainty Estimation} Sampling-based methods derive uncertainty estimation by aggregating multiple outputs. Self-Consistency ~\citep{self-consistency} selects the most frequent answer via majority voting, while Semantic Entropy ~\citep{Semantic-Entropy, SDLG} measures meaning-level consistency across response clusters. For LVLMs, VL-Uncertainty~\citep{zhang2024vl-uncertainty}) estimate uncertainty across multiple samples with semantically equivalent but perturbed inputs.   Internal-state approaches leverage internal signals to quantify certainty. This includes both statistical metrics derived from logits, such as log-probabilities ~\citep{CoCoA}, perplexity ~\citep{UFO-RL}, Self-Certainty (i.e., logits divergence against uniform distribution)~\citep{Self-Certainty}, and the probability assigned to a designated \texttt{true} token~\citep{ptrue}.

\paragraph{Verbalized Confidence Calibration} Recent studies have explored various alignment strategies to calibrate verbalized confidence~\citep{lacie,ppo-c,sayself,rewardingdoubt,RLCR}. One line of work utilizes \textbf{Supervised Fine-Tuning (SFT)} with consistency-based labels ~\citep{sayself} or token-level logit supervision ~\citep{liconftuner}. Another line leverages \textbf{Reinforcement Learning (RL)} to incentivize calibrated uncertainty. For instance, LACIE~\citep {lacie} employs Direct Preference Optimization (DPO)~\citep{dpo,xiao2024comprehensive} within a speaker-listener framework, while PPO-C~\citep{ppo-c}, SaySelf~\citep {sayself}, and Rewarding Doubt~\citep {rewardingdoubt} use Proximal Policy Optimization (PPO)~\citep{ppo} via tailored reward functions (\textit{e.g.}, Brier score or log-penalties) to reward accurate confidence. More recently, RLCR ~\citep{RLCR} uses GRPO~\citep{grpo,xiao2025fastslow} to jointly improve task accuracy and calibration. While effective for text-only tasks, these approaches remain largely unexplored for LVLM calibration.

\section{Methodology}

\label{sec:Methodology}

\subsection{Preliminary}
\label{sec:decoupling}
Let $\pi_\theta$ denote an LVLM, $q = (I, x)$ a multimodal input consisting of an image $I$ and a textual query $x$, $\tau = (z, y)$ a generation trajectory sampled from the policy $\pi_\theta$ conditioned on the multimodal input $(I, x)$, where $z$ denotes the reasoning and $y$ the answer.
Reinforcement Learning with Verifiable Rewards (RLVR; ~\citep{grpo}) optimizes the model using a binary correctness reward:
$R_{\text{acc}}(y, y^*) = \mathbbm{1}_{y \equiv y^*}$, where $\mathbbm{1}_{(\cdot)}$ is indicator function for correctness against the ground-truth $y^*$.

To teach the model to verbalize their confidence alongside the answers, recent work~\citep{RLCR} extends the trajectory to include a holistic confidence score $c \in [0,1]$:
\begin{equation}
    \tau = (z, y, c)
\end{equation}
In addition to correctness reward, it jointly optimizes accuracy and confidence calibration with an additional Brier-based calibration term~\citep{brier}:
\begin{equation}
    R_{\text{acc\_conf}}(y, c, y^*) =
    \mathbbm{1}_{y \equiv y^*} - (c - \mathbbm{1}_{y \equiv y^*})^2
    \label{eq:acc_conf}
\end{equation}

\paragraph{Optimization Objective}
We employ GRPO to optimize $\pi_\theta$.
For each query $q = (I, x)$, we sample a group of $G$ outputs $\{o_1, \dots, o_G\}$ from the old policy $\pi_{\theta_{old}}$.
First, we calculate the composite reward $R_i$ for each output $o_i$ and the advantage $\hat{A}_i$ by normalizing rewards within the group.
The overall objective maximizes the surrogate of $\hat{A}_{i}$ with a KL constraint.
Let $\rho_{i,t}$ denote the token probability ratio between $\pi_\theta$ and  $\pi_{\theta_{old}}$, and $\text{clip}(\cdot)$ denote $\text{clip}(\rho_{i,t}, 1-\epsilon, 1+\epsilon)$.
The objective is defined as:
\begin{equation}
\label{eq:grpo_objective}
\small
\begin{split}
    \mathcal{J}(\theta) &= \mathbb{E}_{q \sim P(Q), \{o_i\} \sim \pi_{\theta_{\text{old}}}} \\
    &\Bigg[ \frac{1}{G} \sum_{i=1}^G \frac{1}{|o_i|} \sum_{t=1}^{|o_i|} \bigg\{\min \bigg( \rho_{i,t} \hat{A}_{i,t},
    \text{clip}(\cdot) \hat{A}_{i,t} \bigg) \\
    &\quad - \beta \mathbb{D}_{KL} \big( \pi_\theta || \pi_{\text{ref}} \big) \bigg\} \Bigg]
\end{split}
\end{equation}
where $\epsilon$ and $\beta$ are the clipping hyperparameter and the coefficient controlling the KL regularization.

\subsection{Decoupling Visual and Reasoning Verbalized Confidence}

To realize verbalized confidence calibration for LVLM reasoning and address the limitations of Eq.~\ref{eq:acc_conf} discussed in ~\S \ref{intro}, we reformulate LVLM reasoning as two phases that explicitly decouples visual perception from reasoning.
Specifically, the policy $\pi_\theta$ is instructed to generate structured visual and reasoning rationales, each followed by a decoupled verbalized confidence:
\begin{equation}
\tau = (\underbrace{z_{vis}, c_{vis}}_{\text{Visual Phase}},
        \underbrace{z_{reas}, c_{reas}}_{\text{Reasoning Phase}}, y)
\end{equation}
where $z_{vis}$ denotes the visual rationale (\textit{e.g.}, image dense caption) and $z_{reas}$ denotes the reasoning chain.
The confidence tokens $c_{vis}, c_{reas}$ represent the model's certainty regarding $z_{vis}$ and $z_{reas}$, respectively, normalized to $\hat{c}_{vis}, \hat{c}_{reas} \in [0, 1]$.

To derive the final holistic confidence $\Phi$ for the answer $y$, we employ the harmonic mean of the decoupled scores:
\begin{equation}
    \Phi(\hat{c}_{vis}, \hat{c}_{reas}) = \frac{2 \cdot \hat{c}_{vis} \cdot \hat{c}_{reas}}{\hat{c}_{vis} + \hat{c}_{reas}}
\end{equation}
We select the harmonic mean for its conservative property: unlike the arithmetic mean, $\Phi$ is dominated by the minimum of the two scores.

To achieve supervision on decoupled visual confidence, we further introduce a Brier-based vision calibration term. Finally, in our decoupled verbalized confidence framework, the reward is denoted as follows:

\begin{equation}
\label{eq:R_tau}
\resizebox{0.95\hsize}{!}{$
\begin{aligned}
R(\tau,y^*, z_{vis}^*) &= \mathbbm{1}_{y \equiv y^*}
- \big(\Phi(\hat{c}_{vis}, \hat{c}_{reas}) - \mathbbm{1}_{y \equiv y^*}\big)^2 \\
&\quad - \big(\hat{c}_{vis} - \mathbbm{1}_{z_{vis} \equiv z_{vis}^*}\big)^2
\end{aligned}
$}
\end{equation}

\subsection{Visual Certainty Estimation}
\label{sec:visual_estimation}
Supervising the decoupled $c_{vis}$ presents a challenge due to the absence of ground-truth labels ($z_{vis^*}$ and $\mathbbm{1}_{z_{vis} \equiv z_{vis}^*}$).
To achieve effective visual certainty estimation as a reliable pseudo label, we propose a composite metric that evaluates visual certainty from two complementary dimensions: \textbf{Visual Grounding} and \textbf{Internal Certainty}.

\paragraph{Visual Grounding.}
Visual grounding measures the extent to which the model's generation relies on the input image rather than hallucinating from language priors~\citep{mathverse,jing2020overcoming}. We quantify this via sensitivity analysis: if an LVLM effectively grounds its rationale in the image, perturbing visual features should significantly alter the output distribution. Conversely, insensitivity implies possible hallucination.
We calculate the KL-divergence between the logits of the original image $I$ and a perturbed version $I'$ via random patch masking with a ratio = 0.8:
\begin{equation}
\small
    D_{KL} = \frac{1}{T} \sum_{t=1}^{T} \text{KL}\left( \pi(\cdot | z_{vis,<t}, I) \parallel \pi(\cdot | z_{vis,<t}, I') \right)
\end{equation}
Here, a high $D_{KL}$ indicates strong grounding, confirming that the generation is actively conditioned on visual tokens.

\paragraph{Internal Certainty.}
However, visual grounding alone is insufficient. A model may rely on the image yet remain conflicted among multiple plausible interpretations (\textit{e.g.}, due to visual ambiguity). To capture this internal state, we measure the average token entropy $\mathcal{H}$ over the visual rationale $z_{vis}$:
\begin{equation}
\small
    \mathcal{H} = -\frac{1}{T} \sum_{t=1}^{T} \sum_{v \in \mathcal{V}} \pi(v | z_{vis,<t}, I) \log \pi(v | z_{vis,<t}, I)
\end{equation}
A low $\mathcal{H}$ indicates a sharp probability distribution, reflecting that the model is internally certain about its generated token.

\paragraph{Visual Certainty Score.}

Intuitively, a high visual certainty score $S_{vis}$ should indicate both \textit{well-grounded} (high $D_{KL}$) and \textit{internally certain} (low $\mathcal{H}$). Therefore, we integrate two metrics to formulate the $S_{vis}$. Considering the diffirent scale of two metrics, we define $S_{vis}$ as a log-ratio signal:
\begin{equation}
    S_{vis} = \log(D_{KL} + \epsilon) - \log(\mathcal{H} + \epsilon)
    \label{eq: S_vis}
\end{equation}
Compared with sampling-based and internal-state estimation methods, including the KL (visual grounding), Entropy (internal certainty), Self-Consistency, Semantic Entropy, VL-Uncertainty, and Self-Certainty, our proposed visual certainty estimation shows superior correlation with \texttt{Gemini-3-pro-preview} perception judgement. Detailed discussion refers to \S \ref{subsec:visual_uncertainty_estimation_validation}.

\subsection{Reinforcement Learning with Certainty-Aware Calibration}
\label{sec:rl}

Building upon the decoupled confidence and visual certainty estimation, we introduce \textbf{Visual Certainty Reward} to provide visual confidence supervision and \textbf{Token-Level Advantage Reweighting} to penalize ungrounded hallucinations while preserving valid visual perception.

\paragraph{Reward Shaping.}
To address the lack of a visual confidence label, we integrate visual certainty estimation via reward shaping. We reformulate the reward function in Eq. \ref{eq:R_tau} as
\begin{equation}
    R(\tau, y^*) = \lambda_{acc} R_{acc} + \lambda_{cal} R_{cal} + \lambda_{vis} R_{vis}
\end{equation}
where $\lambda_{acc}$, $\lambda_{cal}$, and $\lambda_{vis}$ are coefficients. Beyond binary accuracy reward: $R_{acc} = \mathbbm{1}_{y \equiv y^*}$ where $\equiv$ denotes semantical equivalence, and holistic confidence calibration reward: $R_{cal} = - (\Phi(\hat{c}_{vis}, \hat{c}_{reas}) - \mathbbm{1}_{y=y^*})^2$, we propose the \textbf{Visual Certainty Reward} ($R_{vis}$).  Specifically, we use ${S}_{\mathrm{vis}}$ as a proxy of $\mathbbm{1}_{z_{vis} \equiv z_{vis}^*}$. Since the raw visual certainty $S_{vis}$ (Eq.~\ref{eq: S_vis}) varies in scale across batches, we first normalize the raw visual certainty score $S_{\mathrm{vis}}$ by applying batch-wise z-score standardization $z=\frac{S_{\mathrm{vis}}-\mu_B}{\sigma_B+\epsilon}$ and then mapping it to $[0,1]$ via a sigmoid transform $\tilde{S}_{\mathrm{vis}}=\sigma(z)$. Therefore we derive $R_{vis}$ as :
\begin{equation}
    R_{vis} = - \left( \hat{c}_{vis} - \text{sg}(\tilde{S}_{vis}) \right)^2
\end{equation}
where $\text{sg}(\cdot)$ denotes the stop-gradient operator. This term explicitly anchors the model's verbalized visual confidence to its actual perceptual certainty, preventing ungrounded hallucinations.

\paragraph{Token-Level Advantage Reweighting}
Standard GRPO employs uniform credit assignment, treating all errors equally. However, we posit that \textit{ungrounded hallucinations}, where errors arising from high visual uncertainty, indicate a severe perception failure and warrant stricter penalties than other visual errors. To address this, we propose Token-Level Advantage Reweighting (TAR) to dynamically reweight the advantage $\hat{A}_t$ based on the error source and visual certainty. Specifically, we calculate the visual certainty $S_{vis}(t)$ for each specific token. Following the aforementioned Z-score standardization and sigmoid transformation, we normalize these token-wise scores within a single sample to $[0, 1]$ to obtain $\tilde{S}_{vis}(t)$.  We then introduce a reweighting mechanism derived from this token-wise $\tilde{S}_{vis}(t)$ to reweight the advantage of tokens within $z_{vis}$ with $\lambda_{\text{TAR}}= 0.1$ :

\begin{equation}
    \resizebox{.95\hsize}{!}{$
    \hat{A}^{\text{TAR}}_{t} =
    \begin{cases}
    \hat{A}_t \cdot \left( 1 + \lambda_{\text{TAR}} (1 - 2\tilde{S}_{vis}(t)) \right) & \text{if } t \in z_{vis} \land \hat{A}_t < 0 \\
    \hat{A}_t & \text{otherwise}
    \end{cases}
    $}
\end{equation}

Intuitively, when the model errs ($\hat{A}_t < 0$) under low visual certainty ($\tilde{S}_{vis}(t) \to 0$), the penalty is amplified to discourage blind guessing. Conversely, if the model is well-grounded ($\tilde{S}_{vis}(t) \to 1$), the penalty is softened to preserve valid perception.

\begin{table*}[!h]
    \centering

    \definecolor{ourscolor}{gray}{0.93}
    \renewcommand{\arraystretch}{1.1}

    \resizebox{\textwidth}{!}{
    \setlength{\tabcolsep}{2.8pt}
    \begin{tabular}{l @{\extracolsep{4pt}} c c >{\columncolor{ourscolor}}c c c >{\columncolor{ourscolor}}c c c >{\columncolor{ourscolor}}c @{\hspace{10pt}} c c >{\columncolor{ourscolor}}c c c >{\columncolor{ourscolor}}c c c >{\columncolor{ourscolor}}c}
        \toprule
        \multirow{3}{*}{\textbf{Benchmark}} & \multicolumn{9}{c}{\textbf{Qwen3-VL-4B}} & \multicolumn{9}{c}{\textbf{Qwen3-VL-8B}} \\
        \cmidrule(lr){2-10} \cmidrule(lr){11-19}

         & \multicolumn{3}{c}{Accuracy $\uparrow$} & \multicolumn{3}{c}{AUROC $\uparrow$} & \multicolumn{3}{c}{ECE $\downarrow$}
         & \multicolumn{3}{c}{Accuracy $\uparrow$} & \multicolumn{3}{c}{AUROC $\uparrow$} & \multicolumn{3}{c}{ECE $\downarrow$} \\
        \cmidrule(lr){2-4} \cmidrule(lr){5-7} \cmidrule(lr){8-10}
        \cmidrule(lr){11-13} \cmidrule(lr){14-16} \cmidrule(lr){17-19}

         & Base & Best & \textbf{Ours} & Base & Best & \textbf{Ours} & Base & Best & \textbf{Ours}
         & Base & Best & \textbf{Ours} & Base & Best & \textbf{Ours} & Base & Best & \textbf{Ours} \\
        \midrule

        \multicolumn{19}{l}{\textit{\textbf{Mathematical and Geometric Reasoning}}} \\
        DynaMath     & .486 & .718 & \textbf{.753} & .513 & .716 & \textbf{.797} & .423 & .165 & \textbf{.081} & .680 & .766 & \textbf{.784} & .576 & .667 & \textbf{.769} & .460 & .160 & \textbf{.058} \\
        Geo3K        & .514 & .616 & \textbf{.671} & .504 & \textbf{.801} & .792          & .773 & .159 & \textbf{.073} & .514 & .621 & \textbf{.729} & .556 & .761 & \textbf{.780} & .734 & .192 & \textbf{.056} \\
        MathVerse    & .426 & .796 & \textbf{.807} & .416 & .659 & \textbf{.735} & .561 & .142 & \textbf{.042} & .622 & .813 & \textbf{.838} & .504 & .656 & \textbf{.742} & .372 & .129 & \textbf{.055} \\
        MathVision   & .171 & .440 & \textbf{.483} & .501 & \textbf{.814} & .800          & .794 & .207 & \textbf{.170} & .266 & .473 & \textbf{.540} & .527 & .771 & \textbf{.815} & .428 & .249 & \textbf{.094} \\
        MathVista    & .679 & \textbf{.772} & .730 & .566 & .710 & \textbf{.778} & .254 & .132 & \textbf{.107} & .678 & .733 & \textbf{.771} & .574 & .644 & \textbf{.753} & .459 & .198 & \textbf{.079} \\
        WeMath       & .580 & .771 & \textbf{.820} & .593 & .647 & \textbf{.802} & .268 & .164 & \textbf{.048} & .699 & .801 & \textbf{.836} & .567 & .730 & \textbf{.777} & .388 & .110 & \textbf{.039} \\
        \midrule

        \multicolumn{19}{l}{\textit{\textbf{Logical Reasoning}}} \\
        LogicVista   & .456 & .519 & \textbf{.570} & .615 & .757 & \textbf{.794} & .315 & .232 & \textbf{.203} & .508 & .600 & \textbf{.611} & .580 & .688 & \textbf{.836} & .308 & .253 & \textbf{.109} \\
        \midrule

        \multicolumn{19}{l}{\textit{\textbf{Vision-Dominant Reasoning}}} \\
        CLEVR        & .920 & \textbf{.935} & \textbf{.935} & .517 & .577 & \textbf{.797} & \textbf{.025} & .058 & .035 & .910 & .935 & \textbf{.940} & .545 & .495 & \textbf{.723} & .332 & .069 & \textbf{.029} \\
        MathVerse$_V$& .283 & .748 & \textbf{.781} & .519 & .669 & \textbf{.721} & .508 & .171 & \textbf{.056} & .573 & .776 & \textbf{.804} & .502 & .660 & \textbf{.743} & .398 & .162 & \textbf{.052} \\
        \midrule

        \multicolumn{19}{l}{\textit{\textbf{Multi-discipline Reasoning}}} \\
        A-OKVQA      & .836 & .861 & \textbf{.875} & .584 & .592 & \textbf{.695} & .022 & .112 & \textbf{.017} & .829 & .872 & \textbf{.875} & .642 & .593 & \textbf{.691} & .057 & .107 & \textbf{.059} \\
        MMK12        & .489 & .741 & \textbf{.747} & .468 & .651 & \textbf{.714} & .432 & .182 & \textbf{.083} & .585 & .780 & \textbf{.809} & .506 & .691 & \textbf{.777} & .301 & .131 & \textbf{.039} \\
        MMMU-Pro     & .249 & .436 & \textbf{.458} & .610 & .694 & \textbf{.735} & .474 & .340 & \textbf{.335} & .383 & .518 & \textbf{.522} & .579 & .634 & \textbf{.740} & .518 & .357 & \textbf{.220} \\
        ViRL-39K     & .620 & .796 & \textbf{.816} & .406 & .729 & \textbf{.753} & .622 & .113 & \textbf{.026} & .689 & .811 & \textbf{.835} & .537 & .723 & \textbf{.783} & .460 & .109 & \textbf{.033} \\
        \midrule

        \textbf{Avg.} & .516 & .704 & \textbf{.727} & .524 & .694 & \textbf{.763} & .421 & .167 & \textbf{.098} & .610 & .731 & \textbf{.761} & .553 & .670 & \textbf{.764} & .401 & .171 & \textbf{.071} \\
        \bottomrule
    \end{tabular}
    }
    \caption{\textbf{Main Results.} Comparison between the base model using verbalized confidence~\citep{verbalize} (\textbf{Base}), the strongest re-implemented baseline (\textbf{Best}), and our method (\textbf{Ours}) across Qwen3-VL 4B and 8B scales. \textbf{Bold} indicates the best result. Full results of all baselines are reported in Appendix \S~\ref{detailed main results}.}
    \label{tab:main_results}
\end{table*}

\begin{table}[!h]
    \centering
    \small

    \setlength{\tabcolsep}{6pt}
    \begin{tabular}{lcccc}
        \toprule
        \textbf{Model} & \textbf{Method} & \textbf{ACC} & \textbf{AUROC}  & \textbf{ECE}  \\
        \midrule
        \multirow{2}{*}{Qwen3-VL-30B} & P$(\texttt{True})$ & 0.652 & 0.569 & 0.388 \\
        & \cellcolor{gray!10}\textbf{Ours} & \cellcolor{gray!10}\textbf{0.803} & \cellcolor{gray!10}\textbf{0.767} & \cellcolor{gray!10}\textbf{0.082} \\
        \midrule

        \multirow{2}{*}{InternVL3.5-4B} & RLCR & 0.656 & 0.649 & 0.209 \\
        & \cellcolor{gray!10}\textbf{Ours} & \cellcolor{gray!10}\textbf{0.689} & \cellcolor{gray!10}\textbf{0.701} & \cellcolor{gray!10}\textbf{0.103} \\

        \bottomrule
    \end{tabular}
    \caption{\textbf{Generalization Analysis on Different Model Scales and Architectures.} We report the average Accuracy, AUROC, and ECE across all 12 evaluation benchmarks. \textbf{Bold} indicates the best results.}
    \label{tab:generalization}
\end{table}

\section{Experiments}
\subsection{Experimental Setup}

\paragraph{Implementation Details} To control training overhead, we randomly pick 12,000 data points from ViRL-39K~\citep{vl-rethinker}, a diverse categories visual reasoning dataset, as the training dataset, namely VL-Calibration-12K. We apply VL-Calibration on Qwen3-VL-4B-Instruct~\citep{Qwen3-VL}, Qwen3-VL-8B-Instruct, and InternVL3.5-4B-MPO~\citep{wang2025internvl3_5} to confirm efficacy on different model sizes and base models. For more details, refer to Appendix~\ref{appendix: training details}.

\paragraph{Baselines} Our comparison involves inference-stage methods including Verbalize~\citep{verbalize}, P$(\texttt{True})$~\citep{ptrue}, SteerConf~\citep{SteerConf}, and training-stage methods including RLVR~\citep{RLVR}, LACIE~\citep{lacie}, ConfTuner~\citep{liconftuner}, PPO-C~\citep{ppo-c}, SaySelf~\citep{sayself}, Rewarding Doubt~\citep{rewardingdoubt}, and RLCR~\citep{RLCR}. To ensure fair comparisons, we re-implemented these methods on LVLMs with same settings of ours.

\paragraph{Evaluation} To systematically evaluate the efficacy of VL-Calibration, we evaluate on thirteen benchmarks that cover diverse visual reasoning topics, and multi-disciplinary reasoning. Our evaluation metrics include Accuracy (ACC), Expected Calibration Error (ECE), and Area Under the Receiver Operating Characteristic Curve (AUROC) to evaluate models' both reasoning and calibration capability. Details of evaluation benchmarks and metrics are provided in Appendix~\ref {appendix: evaluation details}.

\subsection{Main Results}

Table~\ref{tab:main_results} presents the main results regarding reasoning task performance and calibration. Across both Qwen3-VL-4B and Qwen3-VL-8B models, our method consistently outperforms strong baselines on all metrics. \textbf{First}, our method achieves a significant improvement in ECE, lowering it from 0.421 to 0.098 on the 4B model and from 0.204 to 0.071 on the 8B model.

\textbf{Second}, while existing calibration methods often struggle to maintain the original reasoning accuracy, our method improves the average accuracy by a remarkable margin (\textbf{+2.3\%} over the best baseline on 4B and \textbf{+3.0\%} on 8B), particularly on complex visual reasoning benchmarks like DynaMath and MathVerse.

\textbf{Third}, our method demonstrates out-of-distribution generalization, extending its efficacy to multi-disciplinary tasks. On benchmarks requiring broad knowledge integration like MMMU-Pro, our method consistently outperforms baselines, achieving accuracy gains of +2.2\%. On the commonsense reasoning benchmark A-OKVQA, it reduces ECE from 0.112 to 0.017. We provide significance analysis in Appendix ~\ref{appendix: statistical significance analysis}.

\textbf{Lastly}, to confirm that our method can generalize across various model scales and architectures, we further evaluate its performance on Qwen3-VL-30B and InternVL3.5-4B-MPO. As shown in Table~\ref{tab:generalization}, on the larger 30B scale, it continues to effectively achieve calibration, significantly improving the AUROC to 0.767 and reducing the ECE to 0.082, while simultaneously boosting reasoning accuracy from 0.652 to 0.803. Similarly, when applied to the InternVL architecture, our method outperforms the strong RLCR baseline, achieving a superior task performance (ACC=0.689) and calibration (ECE=0.103).

\subsection{Ablations}

\begin{table}[h]
\centering
\small
\setlength{\tabcolsep}{1.0mm}
\begin{tabular}{l|cc|c|ccc}
\toprule
\multirow{2}{*}{Model} & \multicolumn{2}{c|}{VCE} & \multirow{2}{*}{TAR} & \multicolumn{3}{c}{Metrics} \\
\cmidrule(lr){2-3} \cmidrule(lr){5-7}
 & Ent.& KL & & ACC & AUROC & ECE \\
\midrule
Qwen3-VL-4B & -- & -- & -- & 0.516 & 0.763 & 0.421 \\
RLCR  & -- & -- & -- & 0.704 & 0.694 & 0.167 \\
+ Decoupled & -- & -- & -- & 0.701 & 0.682 & 0.164 \\
\midrule
\multirow{3}{*}{+ VCE} & \checkmark & -- & -- & 0.688 & 0.723 & 0.119 \\
 & - & \checkmark & -- & 0.709 & 0.721 & 0.124 \\
 & \checkmark & \checkmark & -- & 0.715 & 0.751 & 0.121 \\
\midrule
\textbf{Ours} & \checkmark & \checkmark & \checkmark & \textbf{0.727} & \textbf{0.763} & \textbf{0.098} \\
\bottomrule
\end{tabular}
\caption{\textbf{Ablation Study} of Visual Certainty Estimation (VCE) and Token Advantage Reweighting (TAR). Decoupled denotes that we decouple the verbalized confidence in training without VCE and TAR. }
\label{tab:ablation}
\end{table}

We conduct ablation studies to validate the effectiveness of each design component, as presented in Table~\ref{tab:ablation}. \textbf{First, RLCR with decoupling alone does not help calibration}, a variant that only changes the output to $(c_{vis},c_{reas})$ but optimizes the same holistic Brier score performs nearly identically to RLCR, suggesting that explicit visual confidence supervision is necessary. \textbf{Second, on Visual Certainty Estimation (VCE)}, incorporating either entropy or KL-divergence significantly reduces ECE compared to the base model. Notably, the combination of both metrics yields best performance. We empirically observe that relying on a single metric leads to training instability: using only entropy tends to cause \textit{entropy collapse}, while using only KL-divergence risks \textit{entropy explosion}. We provide a detailed visualization and analysis of this phenomenon in \S ~\ref {subsec:combination_analysis}. \textbf{Third, on Token Advantage Reweighting (TAR)}, applying TAR on top of VCE achieves the best overall performance. By uncertainty-aware reweighting the optimization of high-uncertainty tokens advantage, TAR further boosts accuracy to \textbf{0.727} and minimizes ECE to \textbf{0.098}, confirming that fine-grained advantage reweighting is essential for effective calibration.

\section{Analyses}
\label{sec:analyses}

\begin{figure}[t]
    \centering
    \begin{minipage}[t]{1.0\linewidth}
        \centering
        \includegraphics[width=\linewidth]{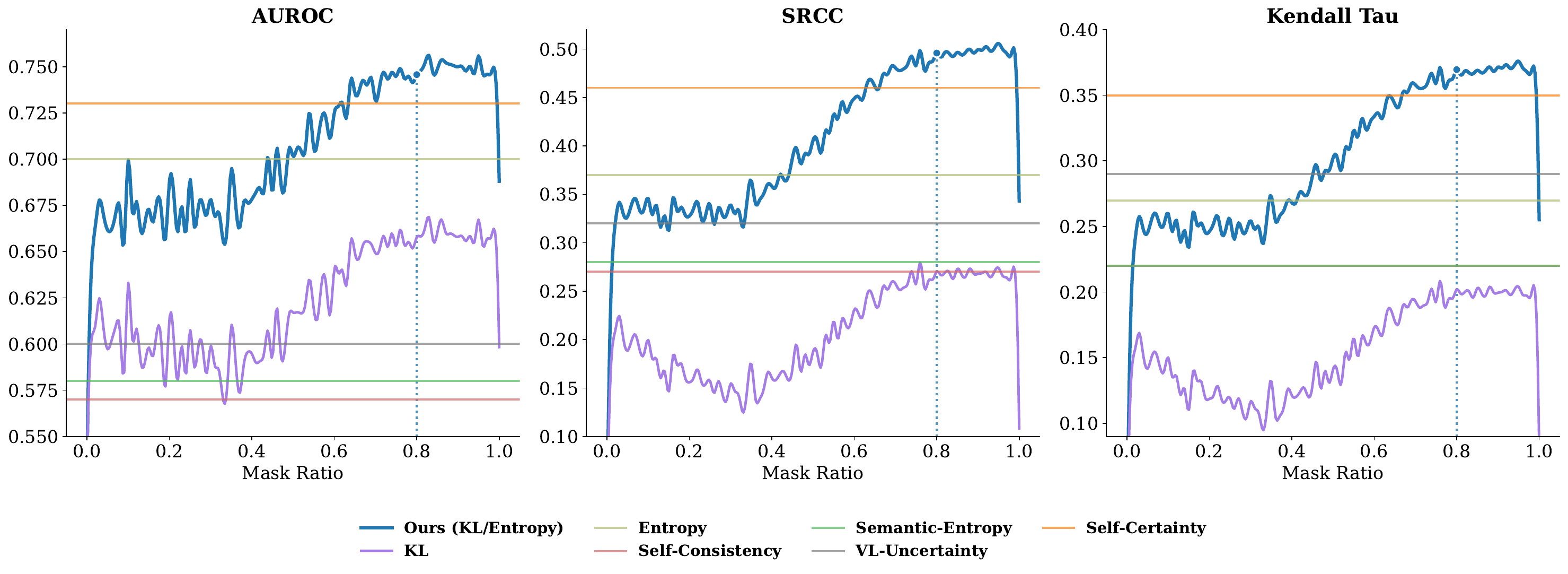}
        \caption{
        \textbf{Effectiveness of Visual Certainty Estimation.}
        Our estimation outperforms the strongest baseline, Self-Certainty, at mask ratios $>0.65$.
        The vertical dashed line marks the mask ratio ($0.8$) adopted in the following experiments.}
        \label{fig: estimation effectiveness}
    \end{minipage}
\end{figure}

\begin{figure}[!h]
    \centering
    \begin{subfigure}[b]{0.32\linewidth}
        \centering
        \includegraphics[width=\linewidth]{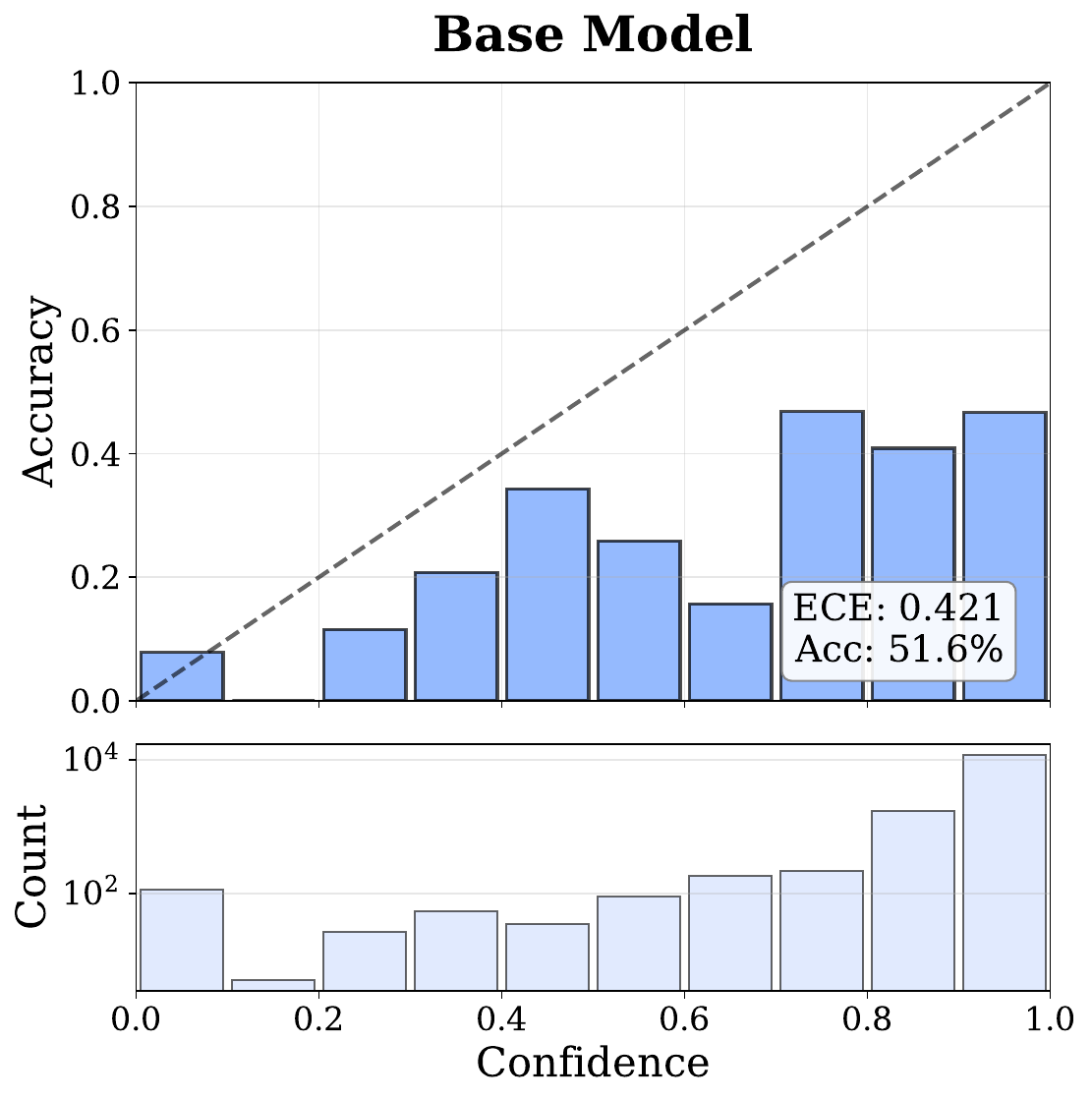}
    \end{subfigure}
    \hfill
    \begin{subfigure}[b]{0.32\linewidth}
        \centering
        \includegraphics[width=\linewidth]{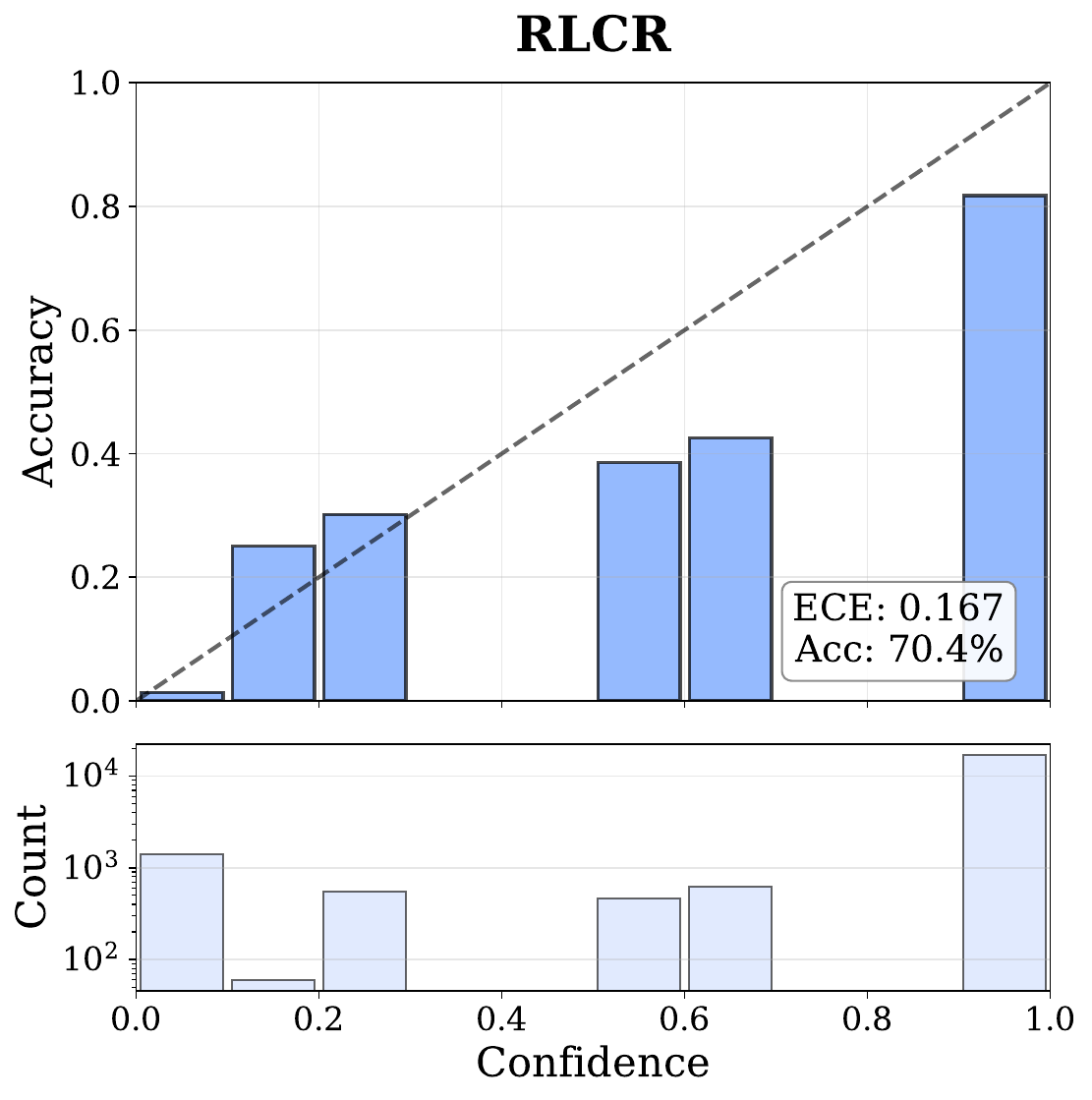}
    \end{subfigure}
    \hfill
    \begin{subfigure}[b]{0.32\linewidth}
        \centering
        \includegraphics[width=\linewidth]{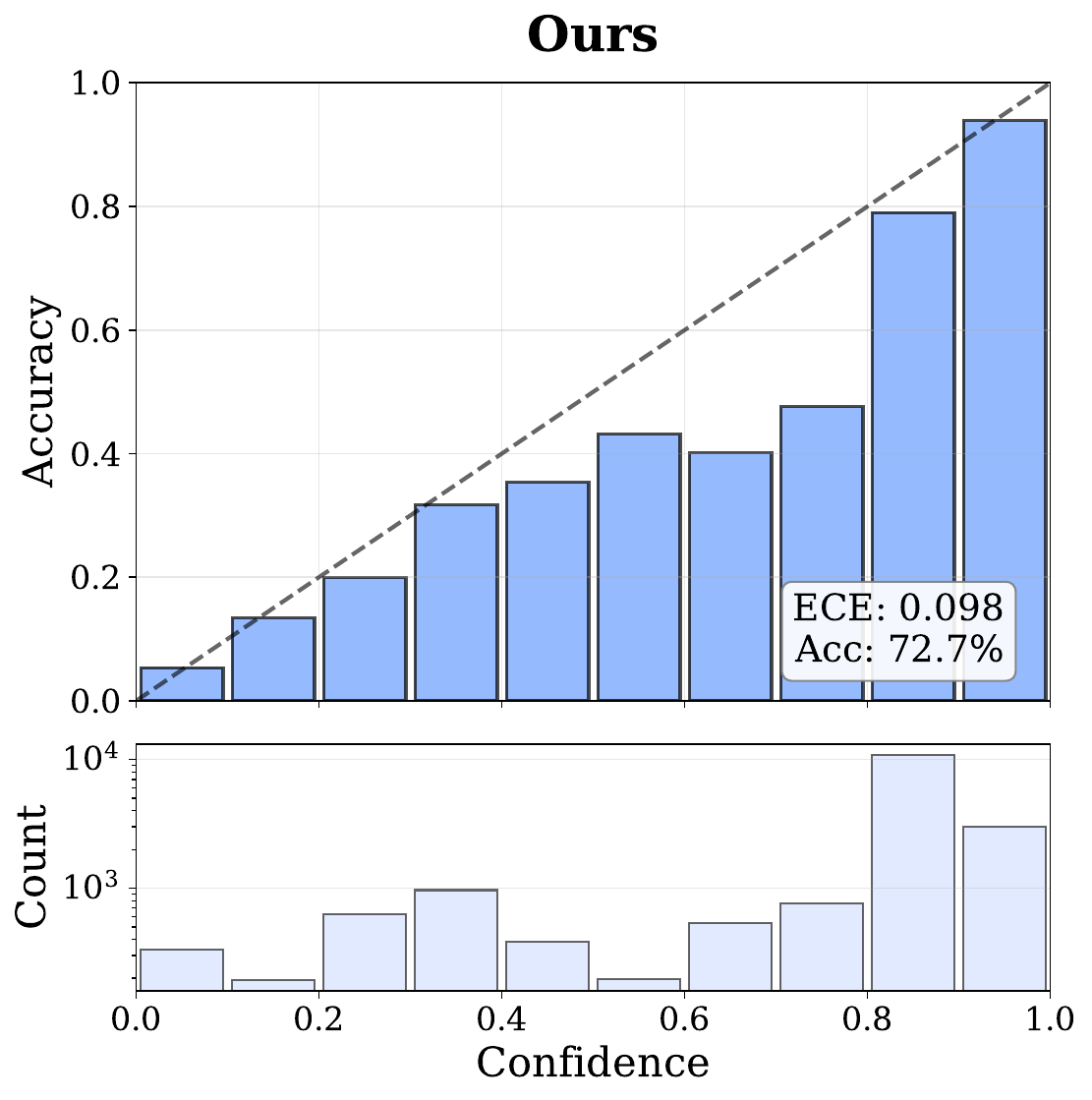}
    \end{subfigure}

    \caption{\textbf{Comparison with Holistic Confidence Calibration.} Reliability diagrams comparison: Base Model (Qwen3-VL-4B, Left), RLCR (Middle), and Ours (Right) across all evaluation datasets.}
    \label{fig: reliablity comparison}
\end{figure}
\paragraph{Validation of Visual Certainty Estimation}
\label{subsec:visual_uncertainty_estimation_validation}
To validate the effect of our visual certainty estimation, we evaluate its correlation with the quality of image captions. Specifically, we use the base model (Qwen3-VL-4B) to generate 1,500 dense captions and employ \texttt{Gemini-3-pro-preview} as a judge to assess the image captions: (i) whether the caption is correct, and (ii) a quality scoring scalar $S_{quality} \in [0, 10]$. Next, we derive uncertainty estimation and utilize AUROC to measure the hallucination detection capability, as well as Spearman’s rank correlation coefficient (SRCC) and Kendall's Tau to evaluate the correlation with quality scores. As illustrated in Figure~\ref{fig: estimation effectiveness}, our proposed estimation surpasses strong baselines, including VL-Uncertainty~\citep{zhang2024vl-uncertainty},  Self-Consistency~\citep{self-consistency}, and Self-Certainty~\citep{Self-Certainty}, with AUROC=0.746, SRCC=0.496, and Kendall's Tau=0.370. Beyond the above validation, we also provide \textbf{more analyses of proposed estimation} regarding computation overhead, combination, and perturbation in Appendix \S ~\ref{appendix: visual_certainty_estimation}.

\paragraph{Training Dynamics}
We present the training dynamics of Qwen3-VL-4B and Qwen3-VL-8B in Figure~\ref{fig:training dynamics}. As illustrated, training with visual certainty estimation exhibits fast initial convergence with ECE reducing to 0.1 in less than 100 steps, achieving higher calibration performance more efficiently. These results indicate that the proposed supervision signal not only improves calibration and reasoning accuracy, but also provides a strong and effective training signal.

\paragraph{Comparison with Holistic Confidence Calibration}
As shown in Figure~\ref{fig: reliablity comparison}, we construct reliability diagrams by binning predicted holistic confidence into $M{=}10$ equal-width bins in $[0,1]$.  For each bin, we plot the average confidence against the empirical accuracy (fraction of correct predictions), with the diagonal line indicating perfect calibration. The reliability diagrams of the base model reveal a severe  \textbf{\textit{overconfidence}} phenomenon, where predicted confidence consistently exceeds accuracy, particularly in high-confidence intervals, with a poor ECE of 0.421, while ours reduce ECE by over $4\times$ to 0.098. Visually, the confidence bins of our model align closely with the diagonal identity line, indicating that the predicted confidence scores serve as a reliable proxy for correctness. Detailed reliability diagrams of each benchmark are provided in Appendix~\ref {appendix: reliability_diagrams}.

\begin{table}[!t]
\centering
\small
\begin{tabular}{lccc}
\toprule
       & \textbf{Unanswerable} & \textbf{Answerable} & $\Delta$ \\
\midrule
Qwen3-VL-4B & 0.698 & 0.926 & 0.228 \\
RLCR & 0.532 & 0.937 & 0.405 \\
\textbf{Ours} & 0.218 & 0.834 & \textbf{0.616} \\
\bottomrule
\end{tabular}
\caption{Comparison of Qwen3-VL-4B, RLCR, and Ours of Confidence Gap ($\Delta$) in visual unanswerable and answerable problems of DynaMath.}
\label{tab:confidence_gap}
\end{table}

\begin{figure}[t]
    \centering
    \begin{subfigure}[b]{\linewidth}
        \centering
        \includegraphics[width=\linewidth]{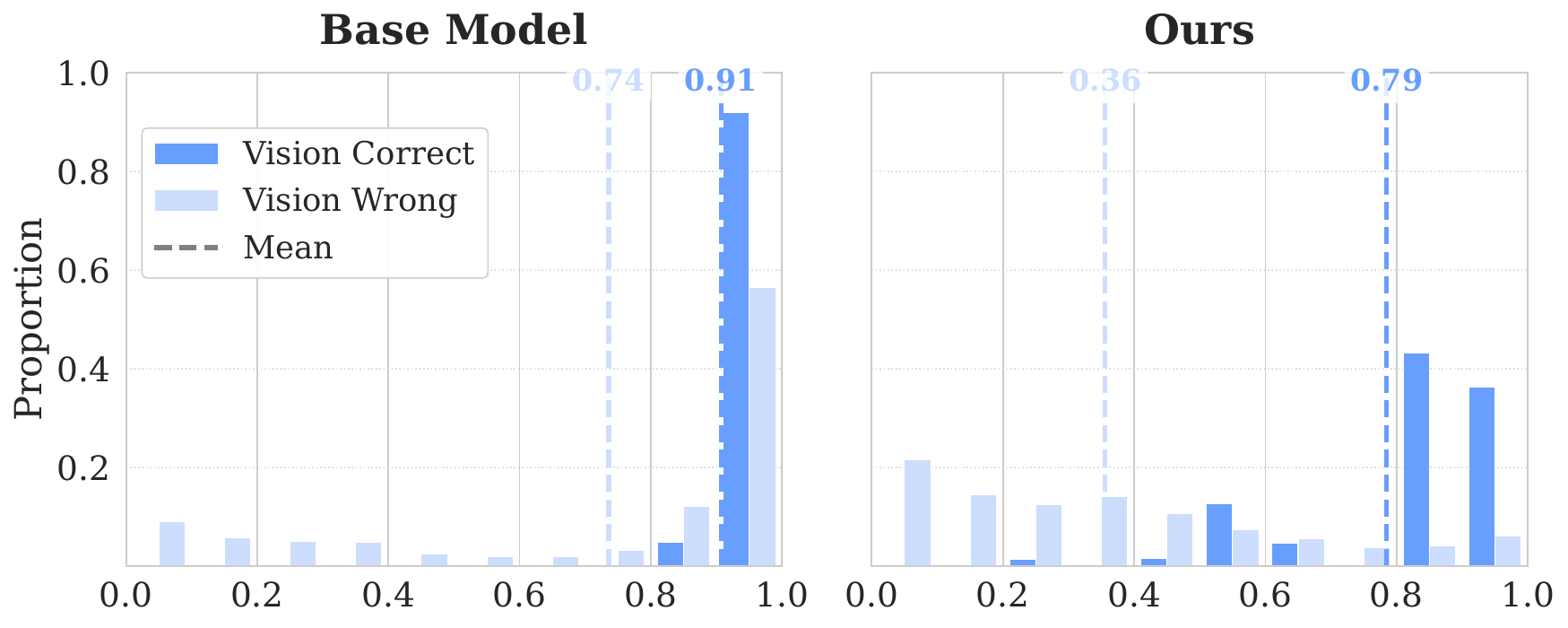}
        \label{fig:visual_conf}
    \end{subfigure}

    \caption{Visual confidence distribution comparison of visually correct and incorrect responses. Base Model: Qwen3-VL-4B. }
    \label{fig:decouple_effectiveness}
\end{figure}

\paragraph{Effect of Proposed Visual Confidence}
To further investigate the effect of decoupled visual confidence, \textbf{first}, we  randomly sample 1,000 problems across all benchmarks to evaluate Qwen3-VL-4B and our model, then manually label responses as visually correct or incorrect. As shown in Figure~\ref{fig:decouple_effectiveness}, the base model assigns high visual confidence to both visually correct and incorrect responses. However, decoupled visual confidence substantially lowers confidence on visually incorrect ones while maintaining relatively high confidence on visually correct ones, effectively distinguishing visual errors. \textbf{Second}, we further evaluate our model's performance on curated \textbf{\textit{visually unanswerable}} problems.
Specifically, we preserve the text input while removing images of \texttt{DynaMath} benchmark, which are generated dynamically to alleviate data contamination.
We report the average confidence on the answerable (original) and visual unanswerable (lack of images) problems in Table~\ref{tab:confidence_gap}.
We observe that, compared to RLCR and the base model, our method achieves the largest \textbf{Confidence Gap ($\Delta$)}, lowering confidence for unanswerable problems while maintaining high certainty for answerable ones.

\begin{figure}
    \centering
    \includegraphics[width=0.98\linewidth]{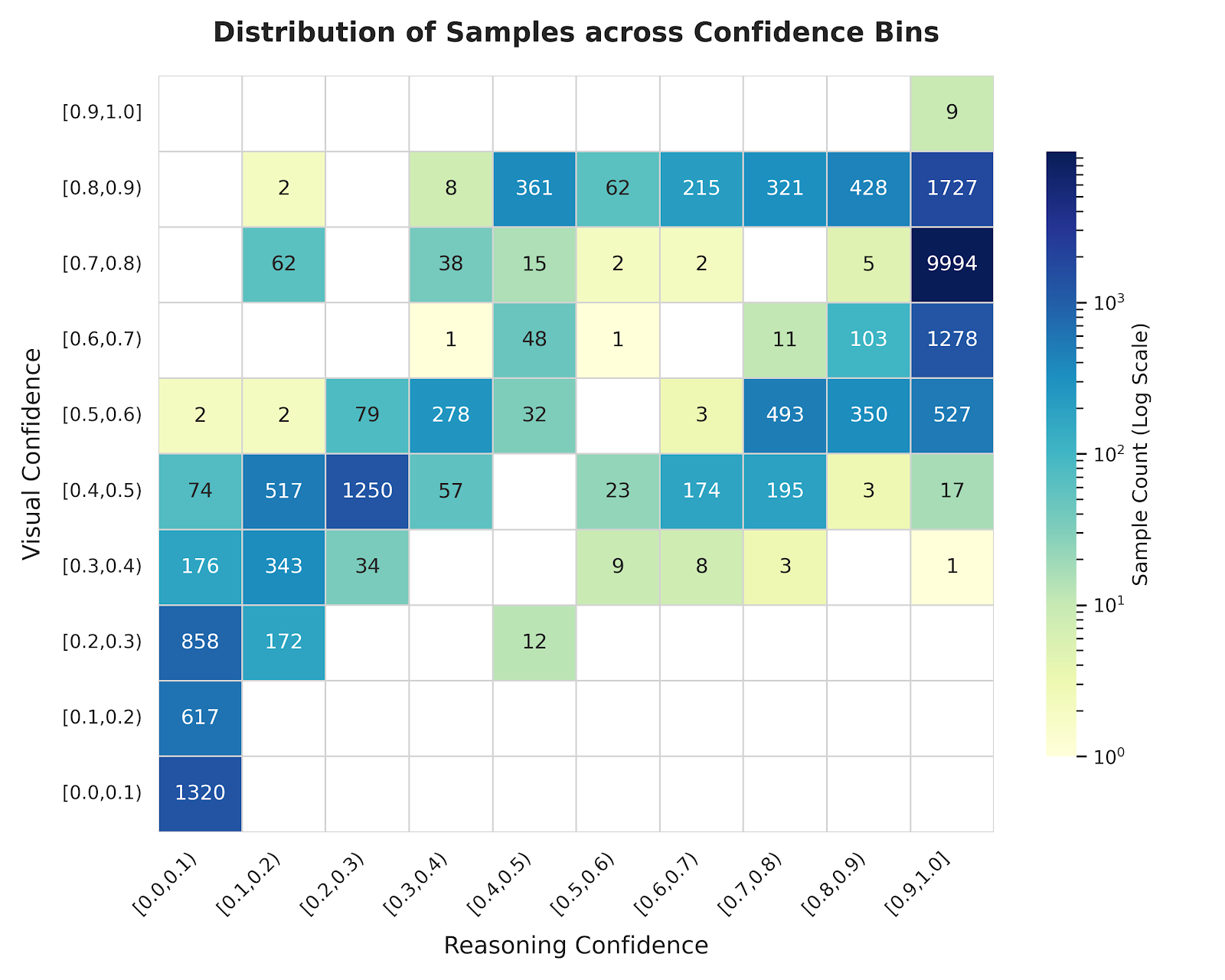}
    \caption{Heatmap of visual vs. reasoning confidence. The off-diagonal distribution indicates that the model can be visually certain but logically uncertain, and vice versa.}
    \label{fig:visual_reasoning_conf}
\end{figure}

\paragraph{Decoupled Distribution of Visual and Reasoning Confidence} To further verify that visual and reasoning confidence measure different things, we visualize their distribution across all samples in a 2D heatmap (Figure ~\ref{fig:visual_reasoning_conf}). The results show that the two scores are clearly separated. For instance, when visual confidence is high (the $[0.8, 0.9)$ bin), reasoning confidence still varies widely from 0.1 to 1.0. Similarly, even when reasoning confidence is at its highest ($[0.9, 1.0]$), visual confidence ranges from 0.3 to 1.0. This indicates that the model can be certain about what it sees but unsure about its logic, or vice versa. This clear separation confirms that a single overall confidence score is insufficient, as it mixes two different sources of uncertainty.

\begin{figure}
    \centering
    \includegraphics[width=0.98\linewidth]{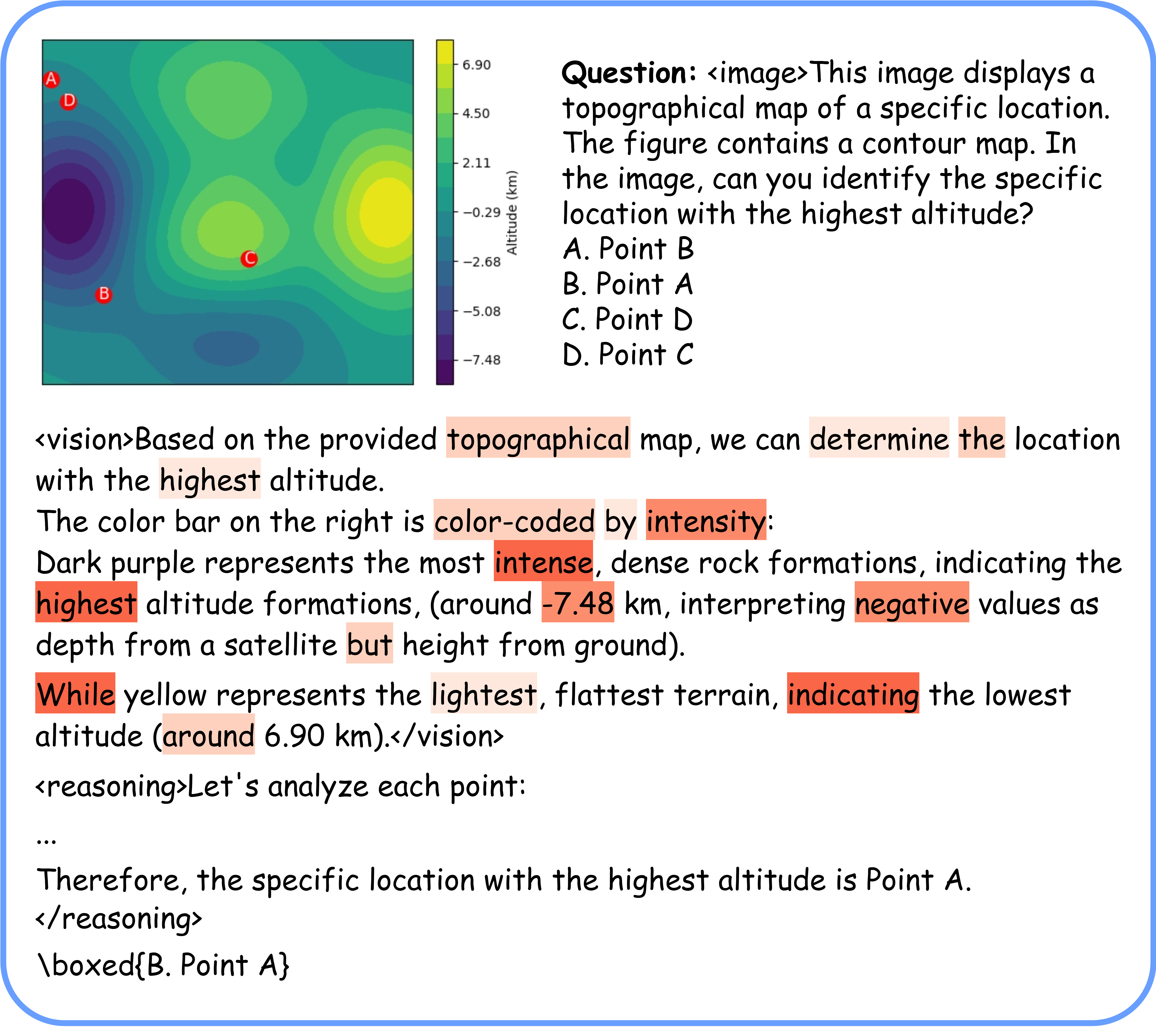}
    \caption{\textbf{Visualization of most visual-uncertain tokens.} Darker red represents higher uncertainty.}
    \label{fig:token_advantage_shaping_sample}
\end{figure}

\paragraph{Qualitative Analysis of Token Advantage Reweighting}
To better understand token advantage reweighting, we visualize the most visually uncertain tokens in Figure~\ref{fig:token_advantage_shaping_sample}. High uncertainty appears not only on visually grounded content tokens (\textit{e.g.}, \texttt{highest}, \texttt{negative}), but also on logical connectives (\textit{e.g.}, \texttt{while} and \texttt{indicating}) that help guide the reasoning process. This observation motivates our use of the visual certainty score for advantage reweighting, where tokens with lower visual certainty receive stronger penalties under negative advantage.

\section{Conclusion}
We present VL-Calibration, an RL-based calibration framework for LVLMs that decouples verbalized confidence into visual and reasoning confidence. We further propose an intrinsic visual certainty estimation signal based on KL-divergence under image perturbation and token entropy, and a token-level advantage reweighting strategy to better suppress ungrounded hallucinations. Experiments on thirteen benchmarks show that VL-Calibration consistently reduces calibration error while improving visual reasoning accuracy, and generalizes across model scales and architectures.

\section{Limitations}
While VL-Calibration demonstrates improved calibration and reasoning across diverse model families and scales, our current evaluation is constrained by computational resources. We present results on Qwen3-VL (4B to 30B) and InternVL3.5-4B, observing consistent gains across these settings. However, the efficacy of our method on larger-scale vision-language models (\textit{e.g.}, 70B+) remains to be empirically verified, as reweighting behaviors may present distinct challenges.

\bibliography{custom}

\clearpage
\newpage
\appendix

\begin{figure*}[t]
    \centering
    \includegraphics[width=\textwidth]{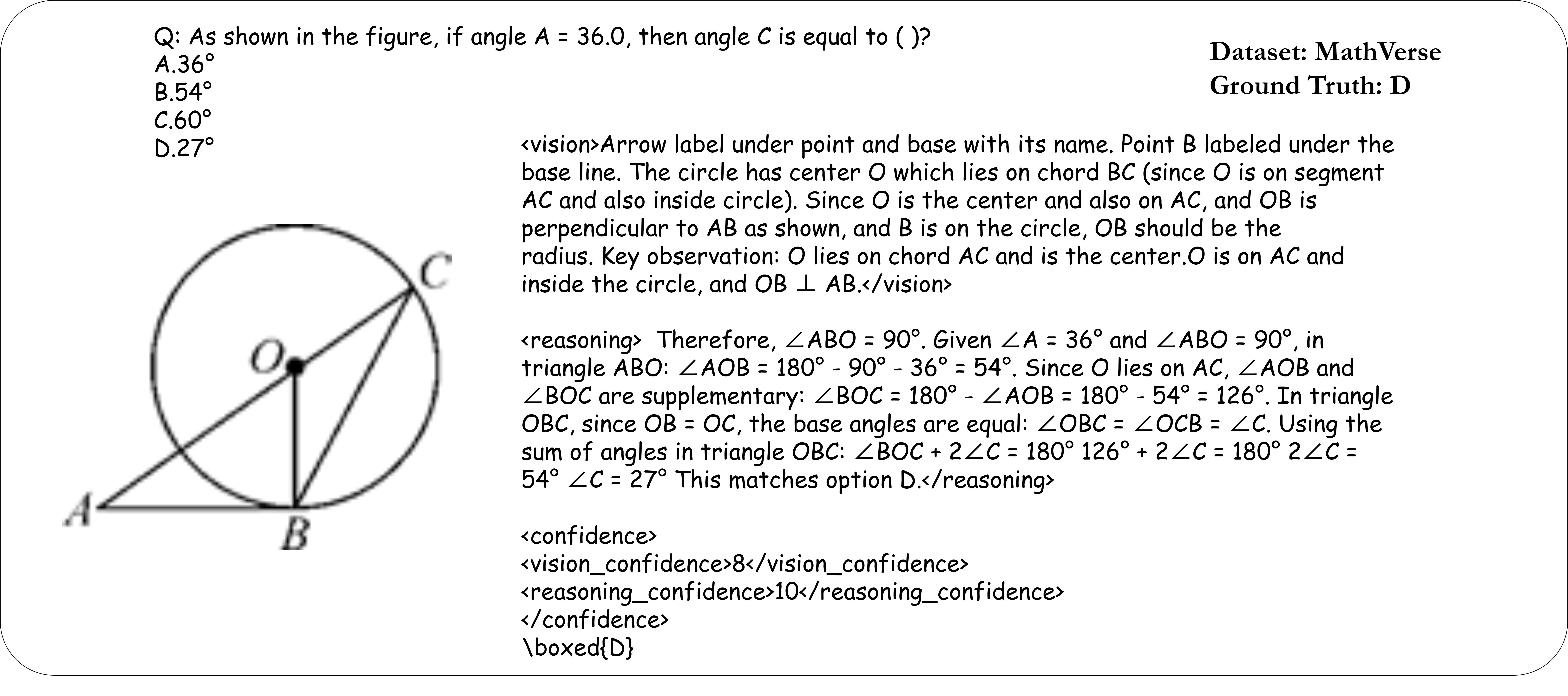}
    \vspace{2mm}
    \includegraphics[width=\textwidth]{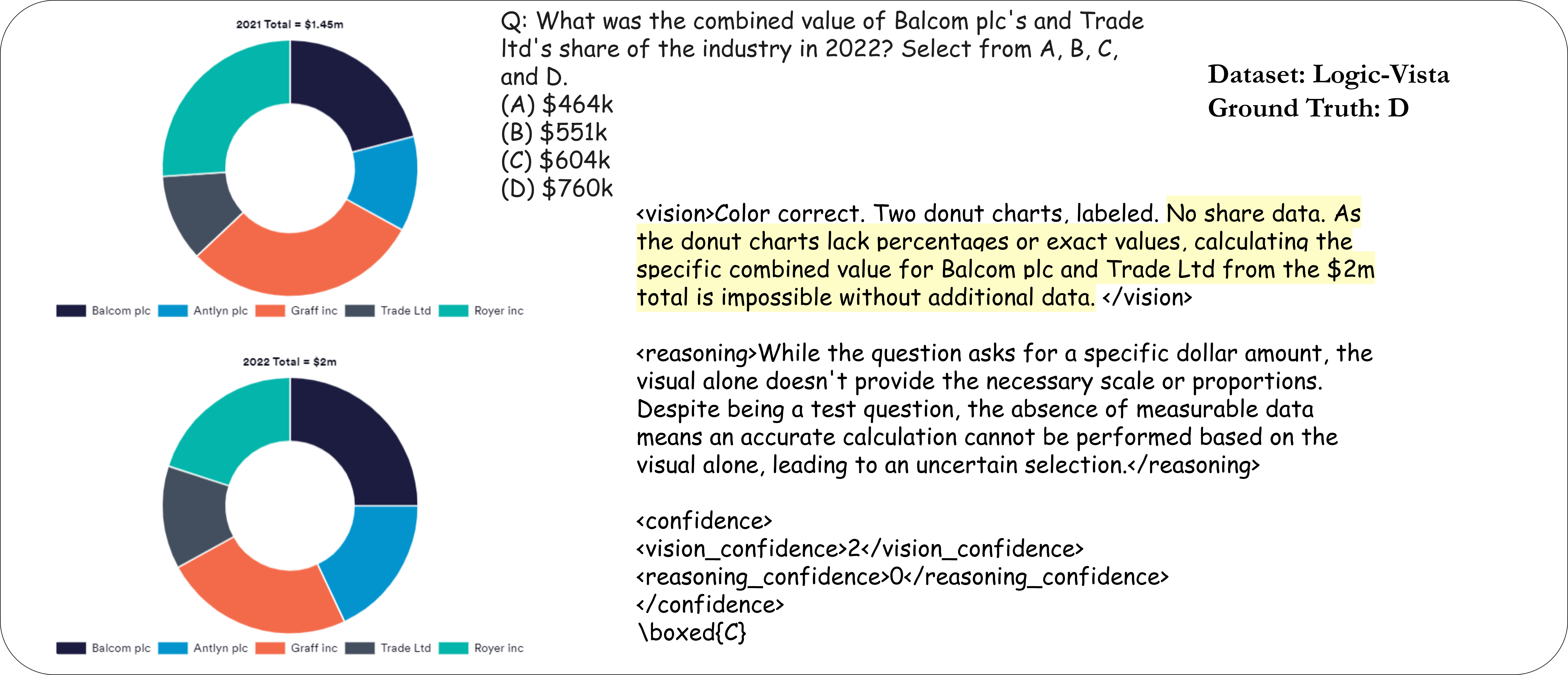}
    \vspace{2mm}
    \includegraphics[width=\textwidth]{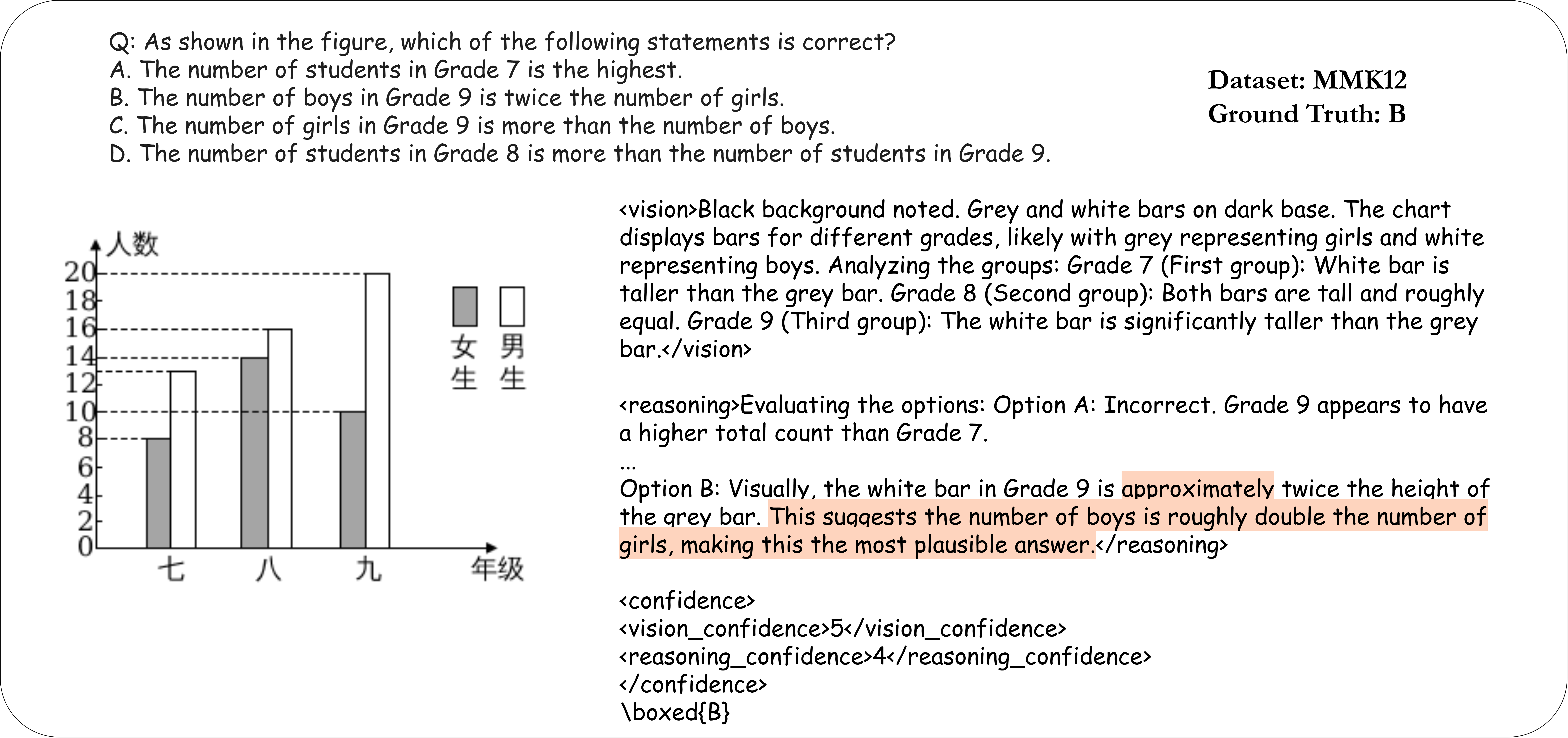}
    \caption{Case Study of Qwen3-VL-4B model trained with our method. Figure (a) (upper) showcases a correct response with high confidence, and Figure (b) (middle) illustrates a incorrect response with low confidence. In Figure (c) demonstrate a correct response with modest confidence.}
    \label{case study}
\end{figure*}

This Appendix for \textit{"VL-Calibration: Decoupled Verbalized Confidence for Large Vision-Language Models Reasoning"} is organized as follows:

\begin{itemize}

    \item \textbf{Experimental Setup and Reproducibility.}
    In \S\ref{appendix: experimental setup}, we describe the experimental setup in detail, including training details (\S\ref{appendix: training details}) and evaluation details (\S\ref{appendix: evaluation details}). The evaluation section further covers the adopted metrics (\S\ref{appendix: evaluation_metrics}) and benchmark datasets (\S\ref{appendix: evaluation_datasets}).

    \item \textbf{Prompts.}
    In \S\ref{appendix: prompt}, we provide the complete \textit{system prompt} used for VL-Calibration training and inference.

    \item \textbf{Statistical Significance Analysis.}
    In \S\ref{appendix: statistical significance analysis}, we report statistical significance analyses to verify the robustness and reliability of the observed performance improvements.

    \item \textbf{Additional Results.}
    In \S\ref{appendix: results}, we present detailed quantitative results, including the complete main results (\S\ref{detailed main results}).

    \item \textbf{Additional Analyses.}
    In \S\ref{appendix: analysis}, we provide more in-depth analyses of VL-Calibration, including visual certainty estimation (\S\ref{appendix: visual_certainty_estimation}), reliability diagrams (\S\ref{appendix: reliability_diagrams}), behavior on unanswerable visual problems (\S\ref{appendix: unanswerable}), training dynamics (\S\ref{appendix: dynamics}), failure mode analysis (\S\ref{appendix: Failure Mode Analysis}), and case study in Figure ~\ref{case study}.
\end{itemize}

\section{Experimental Setup}
\label{appendix: experimental setup}
\subsection{Training Details}
\label{appendix: training details}

\begin{table}[h]
\centering
\caption{Training Hyperparameters}
\label{tab:ppo_hyperparams}
\begin{tabular}{l|c}
    \toprule
        \textbf{Hyperparameter} & \textbf{Value} \\
    \midrule
        Model & Qwen3-VL \\
        Epochs & 15 \\
        Learning Rate & 1e-6 \\
        Train Batch Size & 256 \\
        Temperature & 1.0 \\
        Rollout per Prompt & 8 \\
        Prompt Max Length & 4096 \\
        Generation Max Length & 4096 \\
        Precision & BF16 \\
        Max Pixels & 1000000 \\
        $\lambda_{acc}$ & 1.0 \\
        $\lambda_{cal}$ & 2.0 \\
        $\lambda_{vis}$ & 0.4 \\
        $\lambda_{TAR}$ & 0.1 \\
    \bottomrule
    \end{tabular}
\end{table}

We implement VL-Calibration using Qwen3-VL-4B, 8B, and 30B-A3B as our base models. Below, we detail our training setup and hyperparameters.

\textbf{General Training Hyperparameters.} For VL-Calibration training, we use our 12K dataset with a learning rate of 1e-6, a batch size of 256. We set the maximum sequence length to 4096 for both prompts and generation, and apply BF16 precision throughout training. The training process runs for 15 epochs, requiring approximately 240 H200 GPU hours for Qwen3-VL-4B model, and 450 H200 GPU hours for Qwen3-VL-8B model, 1900 H200 GPU hours for Qwen3-VL-30B-A3B model.

\textbf{Method-specific Training Hyperparameters.} For our reinforcement learning approach, we employ a temperature of 1.0, 8 rollouts per prompt. For the reward weights, we set accuracy reward weight $\lambda_{a} = 1.0$, calibration reward weight
$\lambda_{c} = 2.0$ and visual certainty reward weight $\lambda_{v} = 0.4$. The mask ratio is 0.8.

\textbf{Computation Environment.} All training experiments were conducted using H200 GPUs. Model inference in evaluations is performed using the vLLM framework \citep{vllm}, and our training implementation extends the VeRL codebase \citep{verl}.

The complete set of hyperparameters is provided in Table~\ref{tab:ppo_hyperparams}. We commit to releasing all the code, data, and model checkpoints for experimental results
reproducibility.

\subsection{Evaluation Details}
\label{appendix: evaluation details}

\subsubsection{Evaluation Metrics}
\label{appendix: evaluation_metrics}
We use the following evaluation metrics:

\begin{enumerate}
    \item \textbf{Accuracy:} A measure of reasoning performance.
     \item \textbf{Area Under the Receiver Operating Characteristic Curve (AUROC):} Measures calibration ability of classifier to distinguish between positive/negative classes across thresholds.
    \begin{equation}
        \text{AUROC} = \int_0^1 \text{TPR}(\text{FPR}^{-1}(t)) \, dt
    \end{equation}
     where TPR is the True Positive Rate and FPR is the False Positive Rate.

     \item \textbf{Expected Calibration Error (ECE):} Calibration metric that groups confidences into bins and computes difference between the average correctness and confidence.
    \begin{equation}
        \text{ECE} = \sum_{m=1}^{M} \frac{|B_m|}{N} \left| \text{acc}(B_m) - \text{conf}(B_m) \right|
    \end{equation}
    where $M$ is the number of bins , $B_m$ is the set of samples in bin $m$, and  $N$ is the number of samples.
    We use M=10.
\end{enumerate}

\subsubsection{Evaluation Datasets}
\label{appendix: evaluation_datasets}
This section provides a brief analysis of the eight benchmarks used in our main evaluation. We deliberately selected this suite to cover a wide spectrum of challenges, from domain-specific mathematical skills to general logical cognition, ensuring a holistic assessment of our model's capabilities.

\paragraph{Mathematical and Geometric Reasoning.}
\begin{itemize}
    \item \textbf{DynaMath}~\citep{dynamath} is a unique benchmark designed to test the \textit{robustness} of visual mathematical reasoning. Instead of using a static set of questions, it employs program-based generation to create numerous variants of seed problems, systematically altering numerical values and function graphs to challenge a model's ability to generalize rather than memorize.
    \item \textbf{Geo3k}~\citep{geo3k} is a large-scale benchmark focused on high-school level \textit{geometry}. Its key feature is the dense annotation of problems in a formal language, making it particularly well-suited for evaluating interpretable, symbolic reasoning approaches.
    \item \textbf{MathVerse}~\citep{mathverse} is specifically designed to answer the question: ``Do MLLMs truly see the diagrams?'' It tackles the problem of textual redundancy by providing six distinct versions of each problem, systematically shifting information from the text to the diagram. This allows for a fine-grained analysis of a model's reliance on visual versus textual cues.
    \item \textbf{MathVista}~\citep{mathvista} is a  benchmark designed to combine challenges from diverse mathematical and visual tasks.
    \item \textbf{MATH-Vision}~\citep{mathvision} elevates the difficulty by sourcing its problems from \textit{real math competitions} (e.g., AMC, Math Kangaroo). Spanning 16 mathematical disciplines and 5 difficulty levels, it provides a challenging testbed for evaluating advanced, competition-level multimodal reasoning.

    \item \textbf{We-Math}~\citep{wemath} introduces a novel, human-centric evaluation paradigm. It assesses reasoning by \textit{decomposing composite problems into sub-problems} based on a hierarchy of 67 knowledge concepts. This allows for a fine-grained diagnosis of a model's specific strengths and weaknesses, distinguishing insufficient knowledge from failures in generalization.
\end{itemize}

\paragraph{Logical Reasoning.}
\begin{itemize}
    \item \textbf{LogicVista}~\citep{logicvista} is designed to fill a critical gap by evaluating \textit{general logical cognition} beyond the mathematical domain. It covers five core reasoning skills (inductive, deductive, numerical, spatial, and mechanical) across a variety of visual formats, testing the fundamental reasoning capabilities that underlie many complex tasks.
\end{itemize}

\paragraph{Visual-Dominant Reasoning.}
\begin{itemize}
    \item \textbf{SuperClevr}~\citep{clevr} is a counting benchmark for testing the perception capability.
    \item \textbf{MathVerse$_V$}~\citep{mathverse} We also report MathVerse's vision-dependent subset result, where the problem cannot be solved without its visual input.
\end{itemize}

\paragraph{Multi-discipline Reasoning.}
\begin{itemize}
    \item \textbf{A-OKVQA}~\citep{aokvqa} is a benchmark requiring a broad base of commonsense and world knowledge to answer. The questions generally cannot be answered by simply querying a knowledge base, and instead require some form of commonsense reasoning about the scene depicted in the image.
    \item \textbf{MMK12}~\citep{mmk12} is a benchmark focused on K-12 level multimodal STEM problems. It provides a strong test of foundational scientific reasoning skills that are essential for more advanced applications.
    \item \textbf{MMMU-Pro}~\citep{mmmupro} is a hardened version of the popular MMMU benchmark. It was specifically created to be unsolvable by text-only models by filtering out questions with textual shortcuts, augmenting the number of choices to reduce guessing, and introducing a vision-only format. It serves as a strong test of a model's ability to seamlessly integrate visual and textual information in a high-stakes, academic context.
    \item  \textbf{ViRL-39K-Test}~\citep{vl-rethinker} is a holdout dataset containing 1,800 problems from ViRL-39K excluding VL-Calibration-12K, covering comprehensive topics and categories: from grade school problems to broader STEM and Social topics; reasoning with charts, diagrams, tables, documents, spatial relationships, etc.
\end{itemize}

\section{Prompts}
\label{appendix: prompt}

\begin{minipage}{0.48\textwidth}
\begin{tcolorbox}[colback=gray!5, colframe=gray!40, title=System Prompt]
You FIRST think through the reasoning process as an internal monologue, then provide the final answer.

The reasoning process MUST BE enclosed within <think> </think> tags. Inside the <think> tags, you MUST explicitly separate your thought process into two distinct parts: enclose your visual perception analysis within <vision> </vision> tags, and your logical deduction within <reasoning> </reasoning> tags. The final answer MUST BE put in boxed{}.

After that, perform an <analysis>…</analysis> block to analyse the visual and reasoning confidence in your answer.

Finally, output the confidence scores (0, 1, 2, 3, 4, 5, 6, 7, 8, 9, 10) enclosed within <confidence></confidence> tags. Inside the confidence tags, you MUST strictly output two separate scores enclosed within <visual confidence> </visual confidence> and <reasoning confidence> </reasoning confidence> tags respectively.

\end{tcolorbox}
\end{minipage}

\section{Statistical Significance Analysis}
\label{appendix: statistical significance analysis}
To evaluate the statistical significance of our method on the qwen3-VL-4B model, we conducted 5 independent inference runs with different random seeds. For each run, we recorded the following metrics on the evaluation set: Accuracy, AUROC, and ECE. The average results across the 5 runs are reported in the main paper (ACC = 0.727, AUROC = 0.763, ECE = 0.098). To assess the stability and significance of these results, we computed the mean and standard deviation for each metric as follows:

\begin{table}[h]
    \centering
    \small
    \caption{Stability Analysis across 5 Random Seeds. We report the Mean $\pm$ Standard Deviation.}
    \setlength{\tabcolsep}{4.5pt}
    \label{tab:seed_stability}
    \begin{tabular}{l c c c}
        \toprule
        \textbf{Metric} & \textbf{Acc} ($\uparrow$) & \textbf{AUROC} ($\uparrow$) & \textbf{ECE} ($\downarrow$) \\
        \midrule
        Ours & $0.727 \pm 0.008$ & $0.763 \pm 0.009$ & $0.098 \pm 0.005$ \\
        RLCR & $0.704 \pm 0.007$ & $0.694 \pm 0.010$ & $0.167 \pm 0.006$ \\
        \bottomrule
    \end{tabular}
\end{table}

To confirm whether the improvements over the baseline are statistically significant, we performed paired t-tests on the metrics collected from the 5 independent runs. The significance level was set to $\alpha = 0.05$. The resulting p-values for Accuracy, AUROC, and ECE were $p = 0.012$, $p = 0.008$, and $p = 0.004$, respectively, indicating statistically significant improvements in all three metrics. Overall, these results demonstrate that our proposed decoupled confidence calibration method not only achieves stable performance across different random seeds but also significantly outperforms the baseline method in terms of calibration and accuracy on the Qwen3-VL-4B model.

\section{Results}
\label{appendix: results}

\subsection{Detailed Main Results}
\label{detailed main results}
\begin{figure*}[t]
    \centering
    \includegraphics[width=\textwidth]{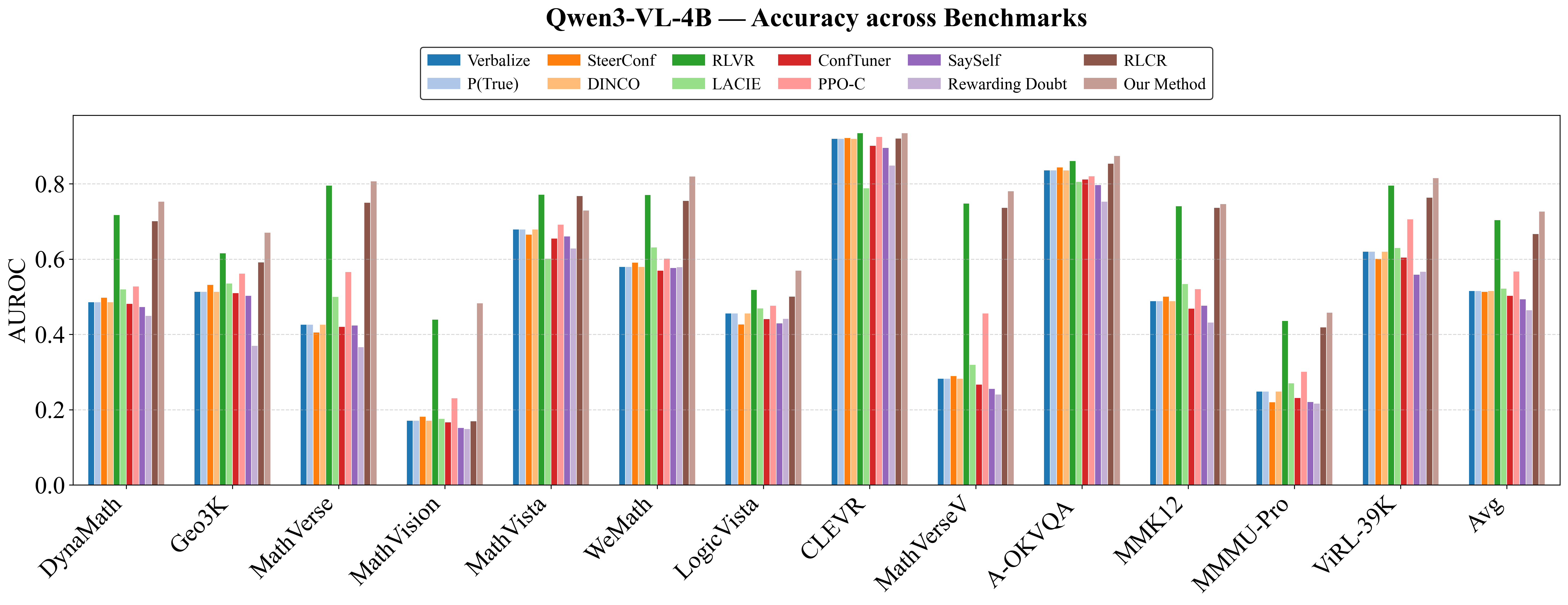}
    \vspace{2mm}
    \includegraphics[width=\textwidth]{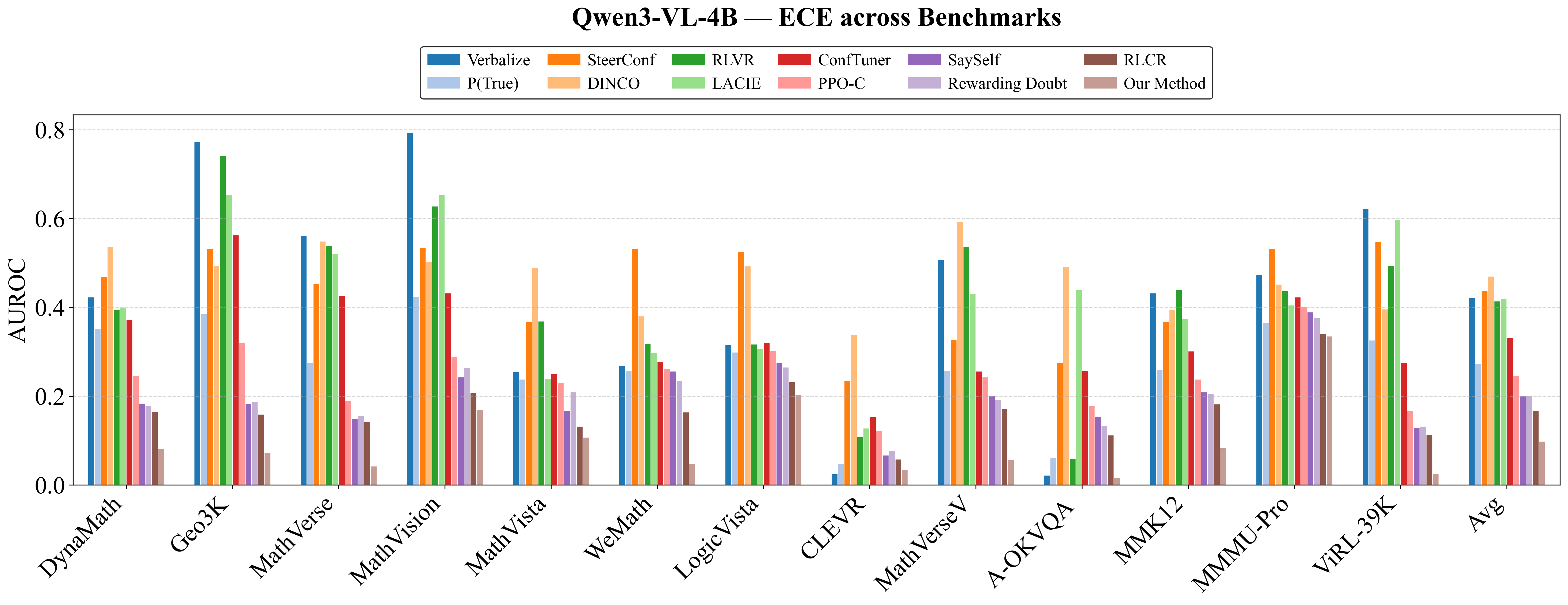}
    \vspace{2mm}
    \includegraphics[width=\textwidth]{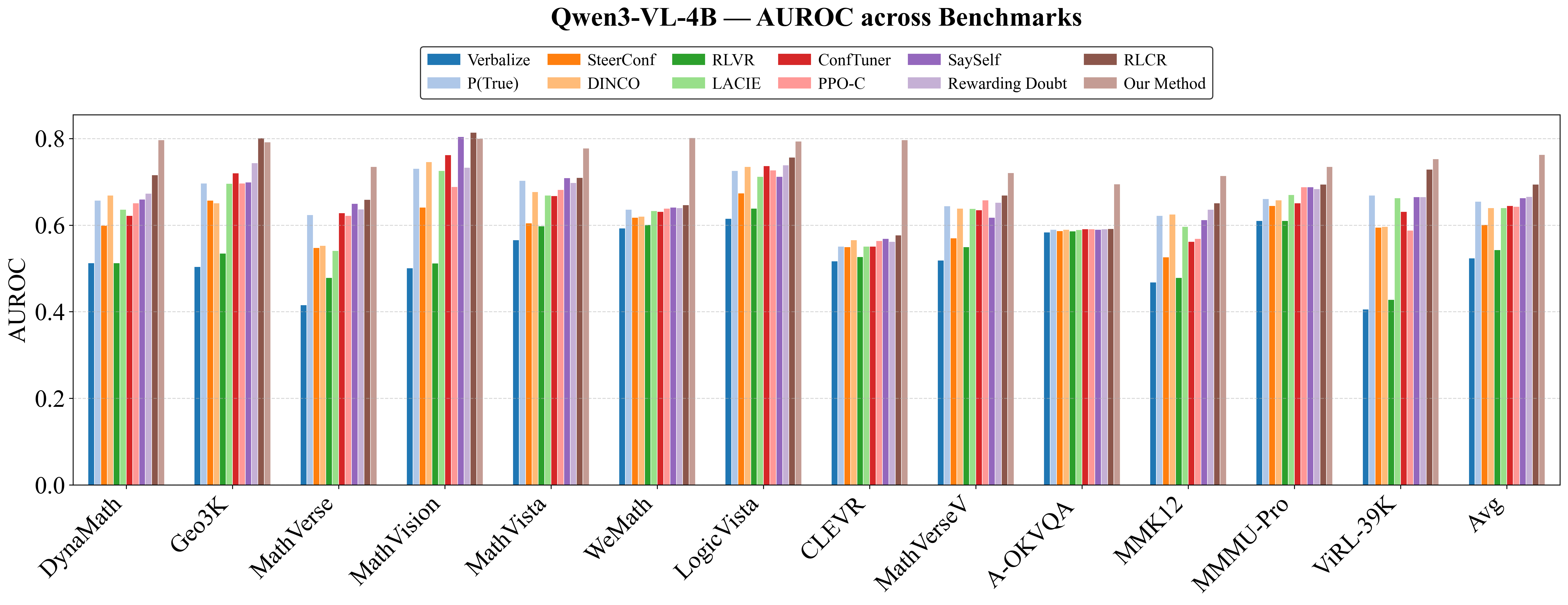}
    \caption{Baselines Performance comparison with Qwen3-VL-4B in terms of Accuracy, ECE, and AUROC.}
    \label{Qwen3-VL-4B baselines}
\end{figure*}

\begin{figure*}[t]
    \centering
    \includegraphics[width=\textwidth]{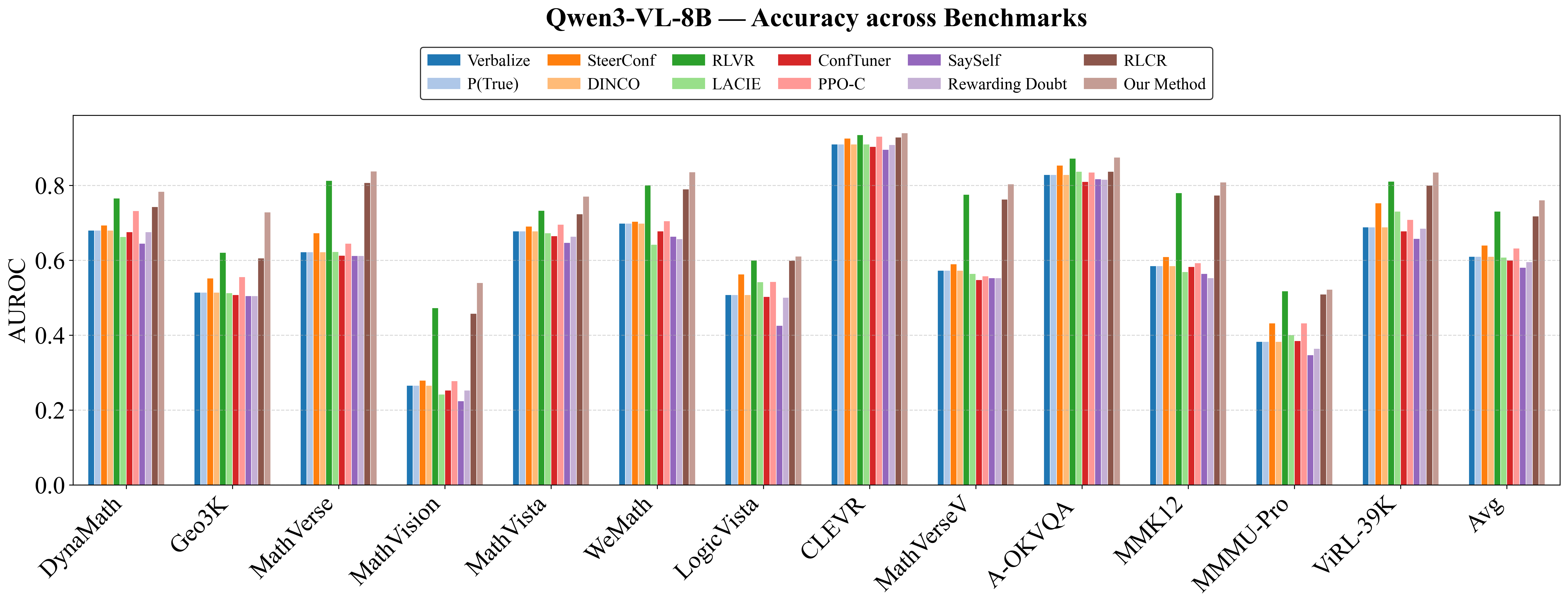}
    \vspace{2mm}
    \includegraphics[width=\textwidth]{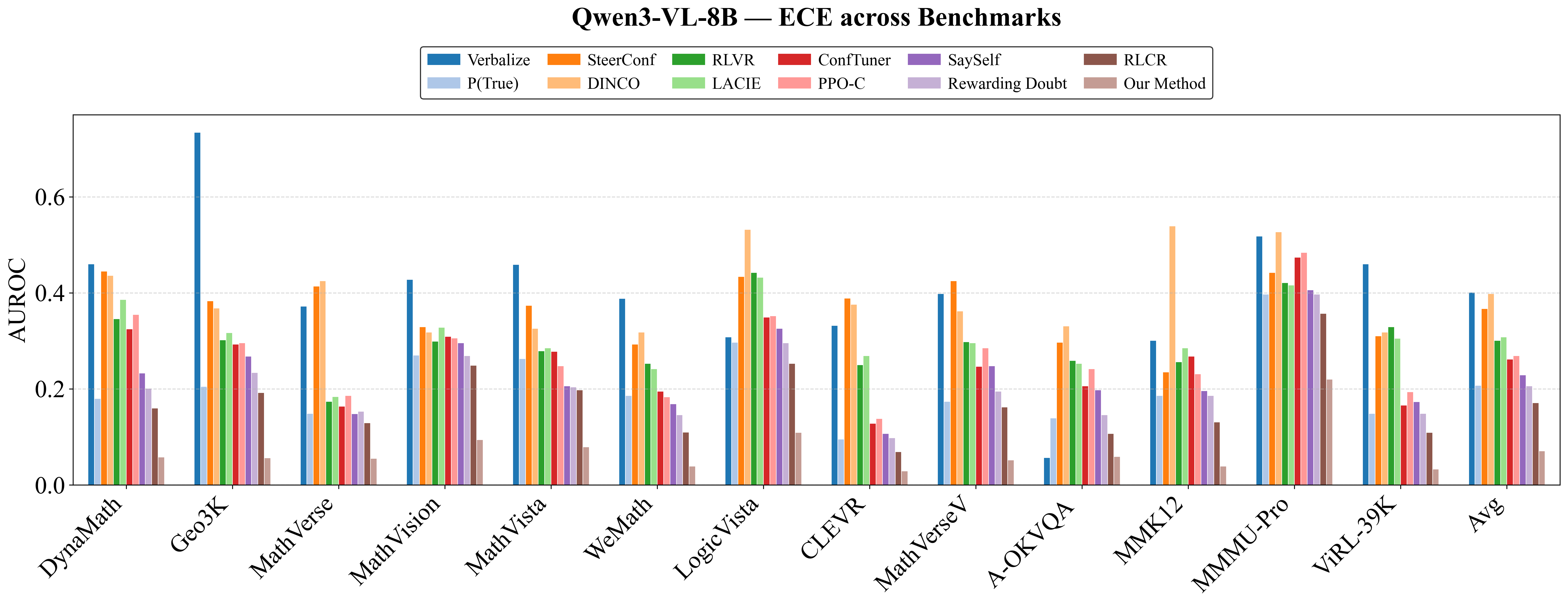}
    \vspace{2mm}
    \includegraphics[width=\textwidth]{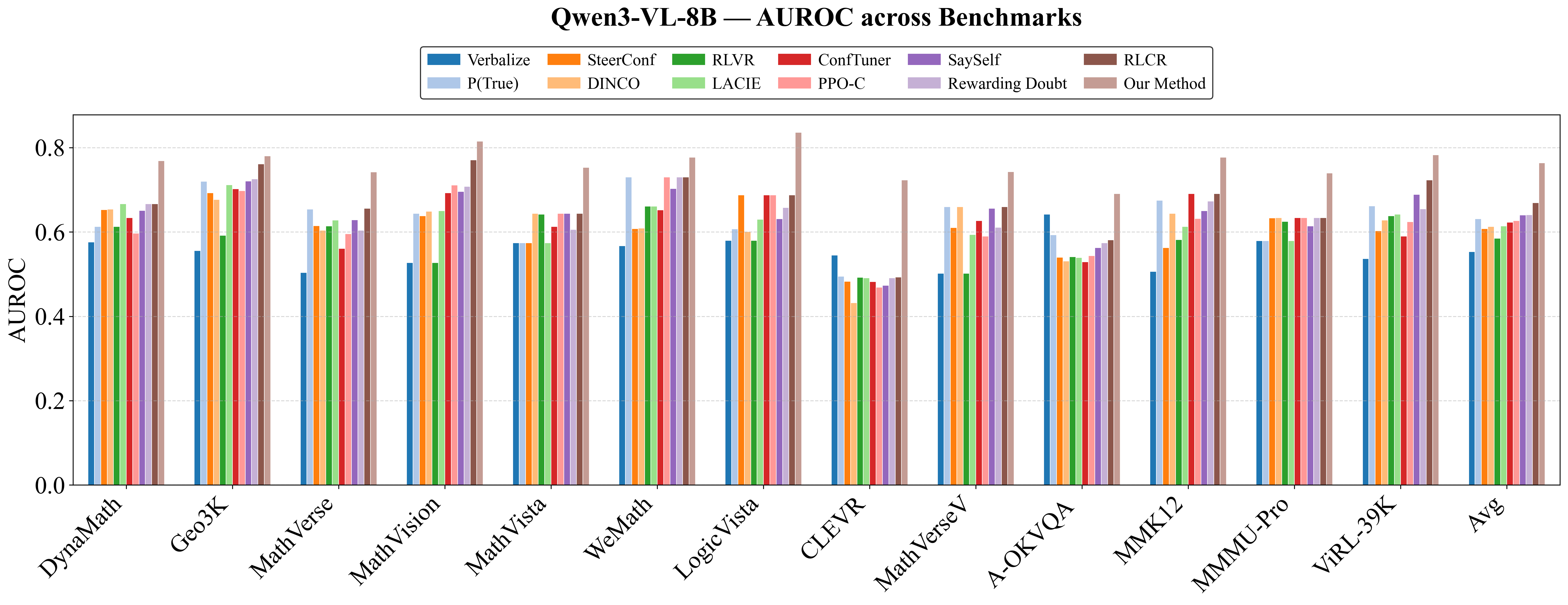}
    \caption{Baselines Performance comparison with Qwen3-VL-8B in terms of Accuracy, ECE, and AUROC.}
    \label{Qwen3-VL-8B baselines}
\end{figure*}

In this section, we report more detailed main results.
We report detailed baseline results on all benchmarks in Figure ~\ref{Qwen3-VL-4B baselines} and Figure~\ref{Qwen3-VL-8B baselines}. We observe that, RLCR is the strongest baseline, outperforming other methods in both reasoning and calibration performance.

\section{Analyses}
\label{appendix: analysis}

\subsection{Visual Certainty Estimation}
\label{appendix: visual_certainty_estimation}

\begin{figure}
    \centering
    \includegraphics[width=1.0\linewidth]{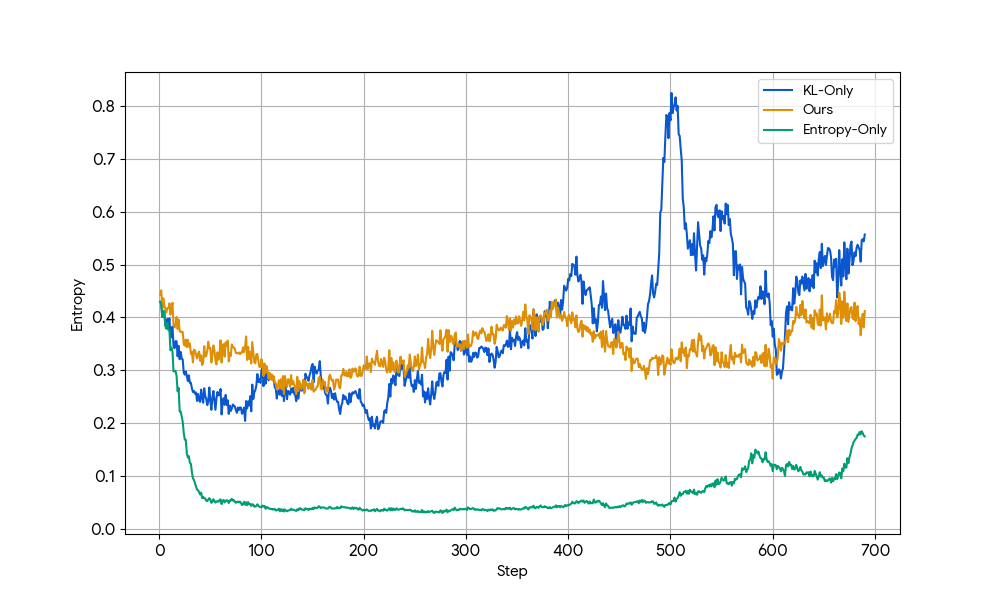}
    \caption{Entropy curves of different estimation: KL-Only, Entropy-Only, Ours (Combination of KL and Entropy). The entropy curve of ours shows training stability compared to others. }
    \label{fig:training_dynamics_kl_entropy}
\end{figure}

\paragraph{Analysis of Certainty Estimation Combination}
\label{subsec:combination_analysis}

Complementing the effectiveness validation in Section~\ref{subsec:visual_uncertainty_estimation_validation}, we further examine the training dynamics in Figure~\ref{fig:training_dynamics_kl_entropy}. We observe that single-metric supervision is prone to optimization pathology: optimizing Entropy only leads to \textit{entropy collapse}, while KL only results in \textit{entropy explosion}. In contrast, our estimation leverages the two metrics as mutual regularizers, effectively preventing both extremes to ensure a robust and stable training trajectory.

\paragraph{Estimation Computation Overhead Analysis}
\label{sec: overhead analysis}
While visual certainty estimation is effective, it introduces additional computational overhead due to the second forward pass required to compute the KL divergence. We analyze this overhead in terms of both runtime and monetary cost, comparing our approach with self-consistency and external annotator baselines.

In terms of runtime, relative to GRPO, the additional forward pass incurs a 11\% time overhead: it adds 15 seconds to the 140-second step time for the 8B model, and 12 seconds to the 100-second step time for the 4B model. By contrast, although self-consistency in GRPO avoids rollout latency, it relies on semantic clustering, which can require up to $O(N^2)$ inferences from an external NLI model in the worst case, leading to substantial latency.

In terms of cost, external annotators are expensive. For example, using \texttt{Gemini-3-pro-preview} costs \$0.03 per judgment, amounting to \$43,200 per training cycle.

\paragraph{Perturbation Design Analysis} To investigate the effect of perturbations on the visual certainty estimation, we evaluate Qwen3-VL-4B across multiple perturbation types and mask ratios. As shown in Table~\ref{tab:perturbation_robustness}, global perturbations (Gaussian Blur and Noise) achieve comparable effectiveness to our default Random Masking (80\%). In contrast, Center Crop underperforms because preserving central objects fails to sufficiently disrupt visual cues.

\begin{table}[htbp]
\centering
\caption{Robustness evaluation of Qwen3-VL-4B across different perturbation types and severities.}
\label{tab:perturbation_robustness}
\resizebox{1.00\columnwidth}{!}{
\begin{tabular}{llccc}
\toprule
\textbf{Perturbation Type} & \textbf{Settings} & \textbf{Accuracy} ($\uparrow$) & \textbf{AUROC} ($\uparrow$) & \textbf{ECE} ($\downarrow$) \\
\midrule
\textbf{Random Masking (Ours)} & \textbf{Ratio = 80\%} & \textbf{0.727} & \textbf{0.763} & \textbf{0.098} \\
Random Masking & Ratio = 20\% & 0.650 & 0.688 & 0.188 \\
Random Masking & Ratio = 50\% & 0.682 & 0.711 & 0.151 \\
Random Masking & Ratio = 100\% & 0.691 & 0.722 & 0.142 \\
Gaussian Blur & $\sigma = 5.0$ & 0.708 & 0.758 & 0.105 \\
Gaussian Noise & $\sigma = 0.5$ & 0.714 & 0.749 & 0.110 \\
Center Crop & Crop 50\% & 0.699 & 0.706 & 0.148 \\
\bottomrule
\end{tabular}
}
\end{table}

Regarding mask ratio, performance improves as the ratio increases from 20\% to 80\%, since weaker masks leave excessive visual information that trivializes the grounding evaluation. However, complete masking (100\%) slightly degrades performance, aligning with Figure ~\ref{fig: estimation effectiveness} where optimal certainty estimation peaks at a 0.8 ratio. Overall, these results confirm that our metric is robust across perturbation designs.

\paragraph{Aggregation Function Analysis}
To justify using the Harmonic Mean for aggregating visual ($c_{vis}$) and reasoning ($c_{reas}$) confidences, we compare it against alternative aggregation functions.

\begin{table}[htbp]
\centering
\caption{Comparison of different aggregation functions for decoupled confidence on Qwen3-VL-4B.}
\label{tab:aggregation_ablation}
\resizebox{1.00\columnwidth}{!}{
\begin{tabular}{lccc}
\toprule
\textbf{Aggregation Function} & \textbf{Accuracy} ($\uparrow$) & \textbf{AUROC} ($\uparrow$) & \textbf{ECE} ($\downarrow$) \\
\midrule
Arithmetic Mean & 0.725 & 0.741 & 0.145 \\
Geometric Mean & 0.724 & 0.752 & 0.107 \\
Minimum & 0.718 & 0.749 & 0.121 \\
\textbf{Harmonic Mean} & \textbf{0.727} & \textbf{0.763} & \textbf{0.098} \\
\bottomrule
\end{tabular}
}
\end{table}
Technically, a reliable aggregated confidence should only be high when both decoupled confidences are high. The \textit{Arithmetic Mean} is overly optimistic; for example, a severe visual hallucination ($c_{vis} \approx 0$) coupled with strong language priors ($c_{reas} \approx 1$) still yields an aggregated score of $0.5$. The \textit{Minimum} function is strictly conservative but suffers from vanishing reward signals for the non-minimum term, which hinders effective joint optimization. Compared to the \textit{Geometric Mean}, the \textit{Harmonic Mean} serves as a more conservative penalty.

Empirically, we evaluate these functions during the RL training of Qwen3-VL-4B. As shown in Table~\ref{tab:aggregation_ablation}, the Harmonic Mean achieves the optimal balance, yielding the highest accuracy and AUROC, along with the lowest ECE.

\subsection{Reliability Diagrams}
\label{appendix: reliability_diagrams}
\begin{figure*}[!htbp]
 \centering
 \includegraphics[width=\textwidth,height=\textheight,keepaspectratio]{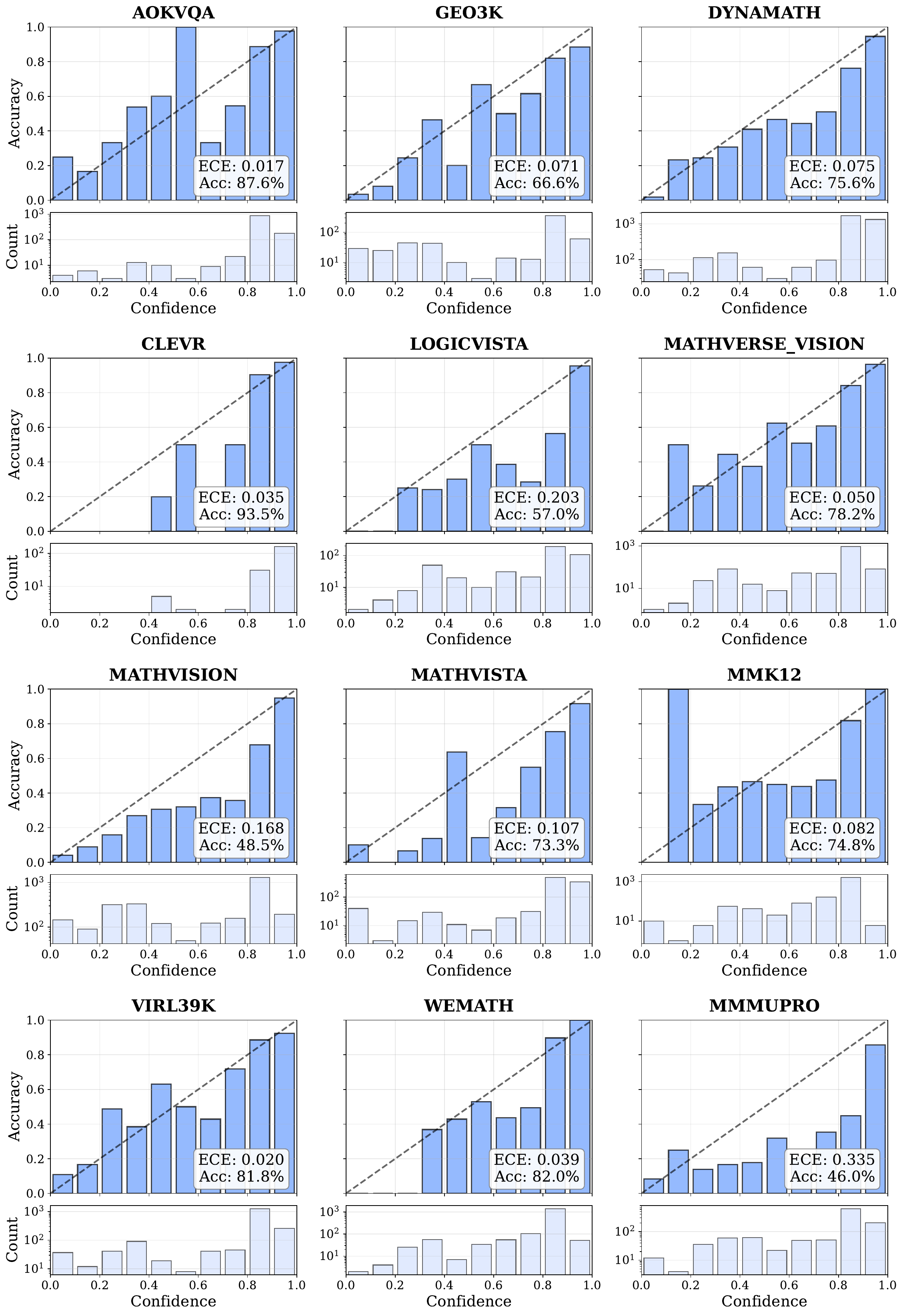}
    \caption{Detailed reliability diagrams of VL-Calibration-4B across the evaluation datasets. While Table~\ref{tab:comparison_grouped} reports metrics averaged over 8 rollouts (avg@8), this figure illustrates the results of a single representative rollout.}
 \label{fig:full_page_reliability_diagrams_4b}
\end{figure*}

\begin{figure*}[!htbp]
 \centering
 \includegraphics[width=\textwidth,height=\textheight,keepaspectratio]{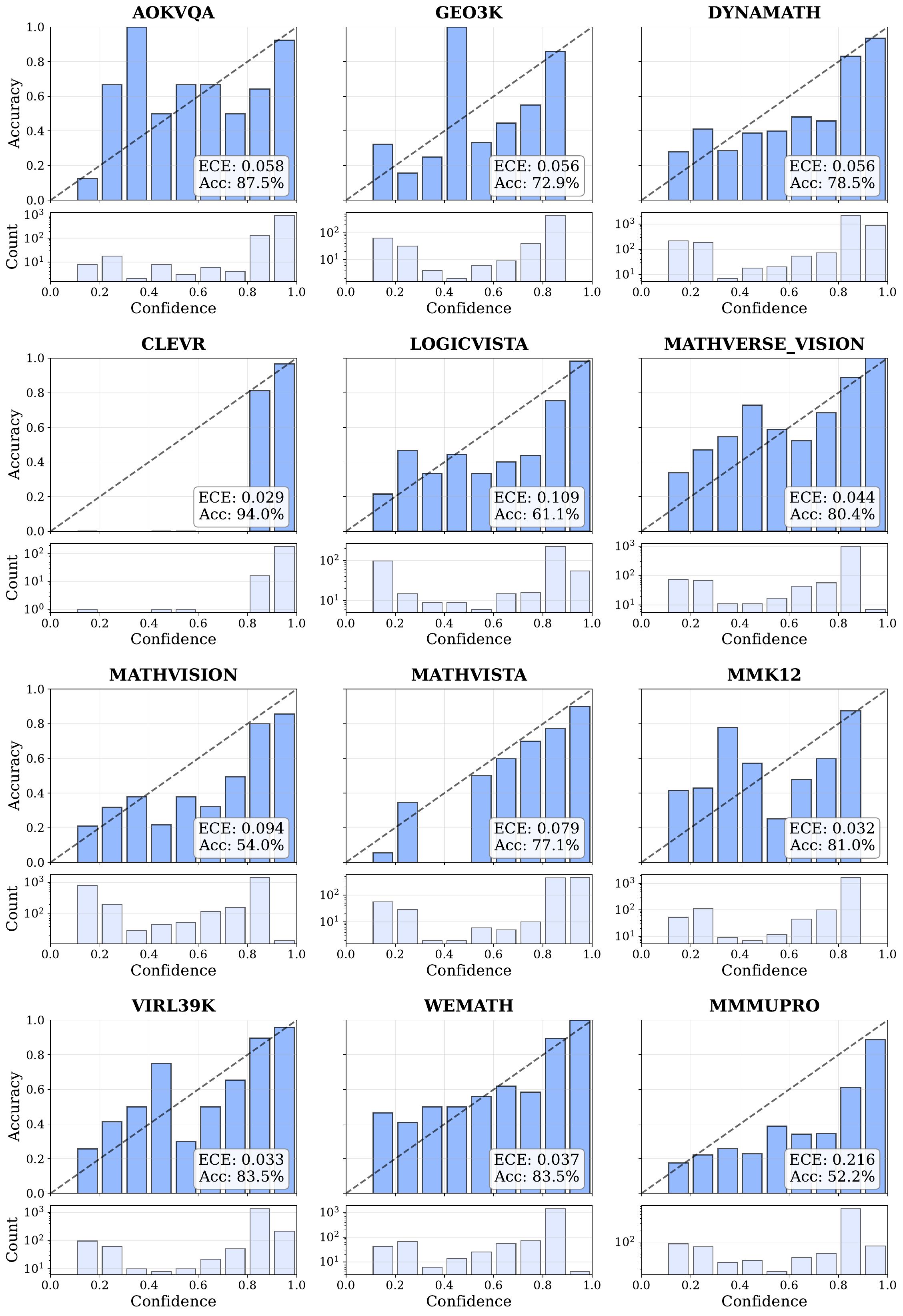}
    \caption{Detailed reliability diagrams of VL-Calibration-8B across the evaluation datasets.}
 \label{fig:full_page_reliability_diagrams_8b}
\end{figure*}

We illustrate the reliability diagrams of VL-Calibration-4B and VL-Calibration-8B on each benchmark in Figure~\ref {fig:full_page_reliability_diagrams_4b} and Figure~\ref{fig:full_page_reliability_diagrams_8b}.

\subsection{Visually Unanswerable Problems}
\label{appendix: unanswerable}

\begin{figure*}[t]
  \centering
  \includegraphics[width=\textwidth]{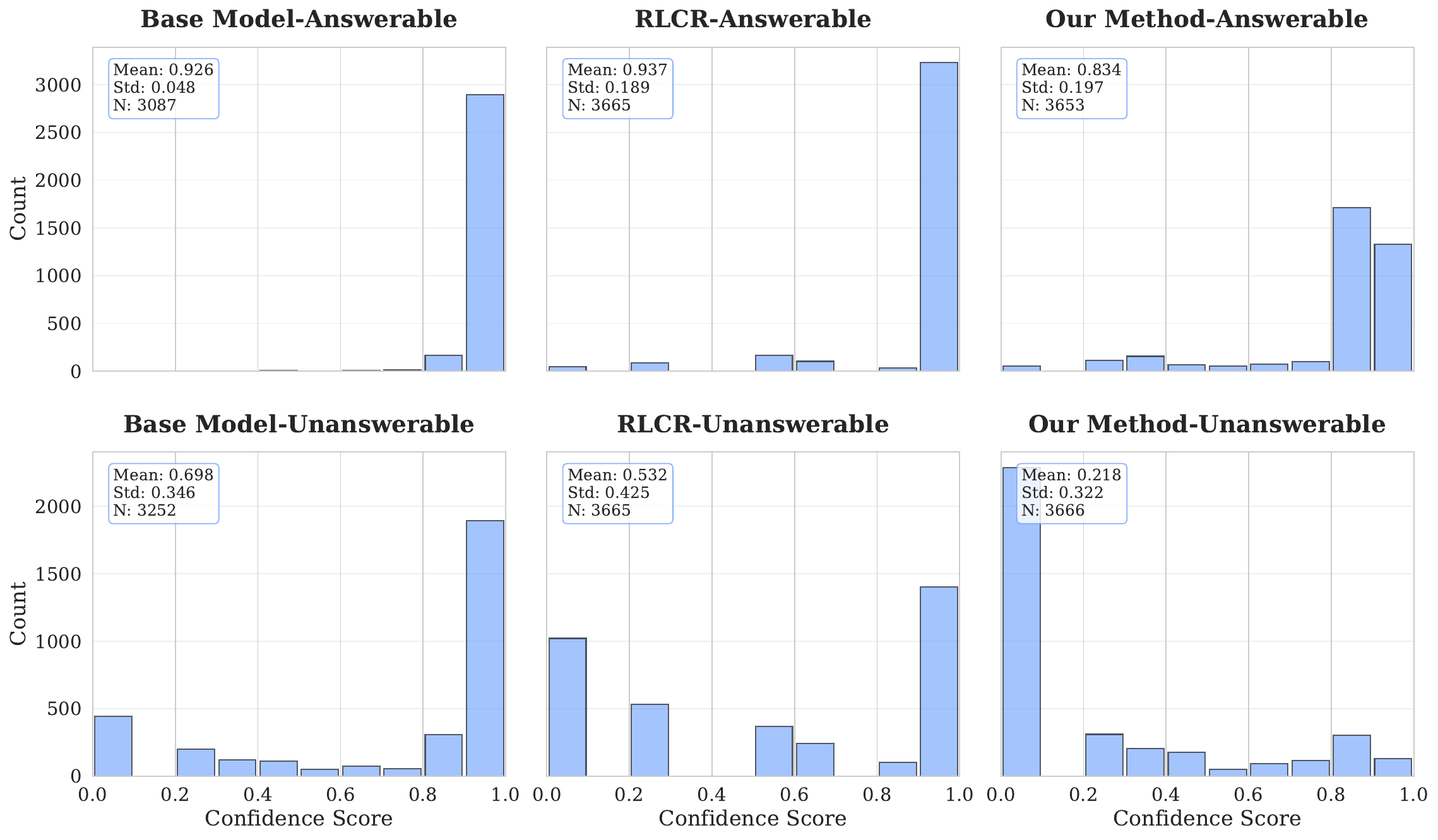}
    \caption{
    \textbf{Confidence Distribution across Visually Answerable and Unanswerable Problems.}
    We visualize the distribution of confidence scores for the Base Model, RLCR, and Our Method (columns) on both Answerable and Unanswerable datasets (rows). While baselines tend to remain overconfident even on unanswerable queries (bottom row), \textbf{Our Method} exhibits a significant distributional shift towards lower confidence, demonstrating superior capability in recognizing visual input lack.
    }
  \label{fig:confidence_distribution_answerable}
\end{figure*}

In Figure~\ref{fig:confidence_distribution_answerable}, we provide concrete confidence distributions of Base Model, RLCR, and Ours across visually answerable and unanswerable problems.

\subsection{Training Dynamics}
\label{appendix: dynamics}

\begin{figure*}[!h]
  \centering
    \includegraphics[width=\textwidth]{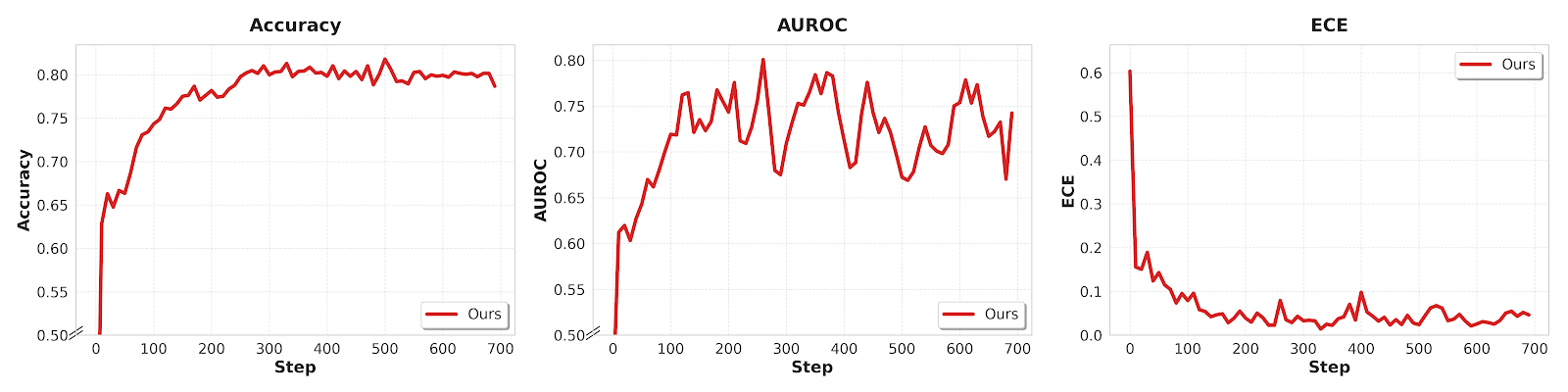}
  \includegraphics[width=\textwidth]{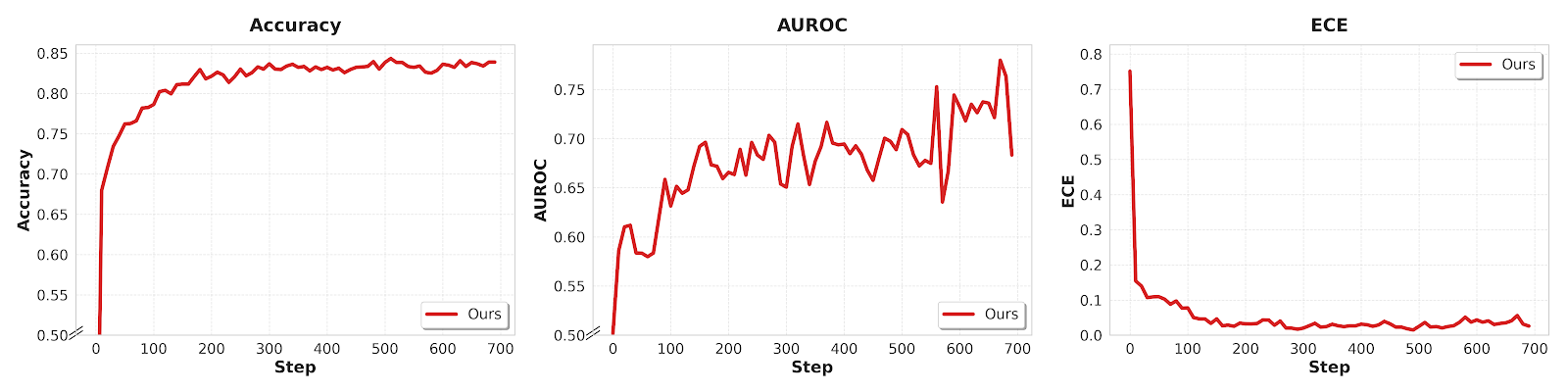}

    \caption{Training dynamics of Qwen3-VL-4B (upper) and Qwen3-VL-8B (bottom). We visualize the ACC, AUROC, ECE curves on the validation set. }
  \label{fig:training dynamics}
\end{figure*}

We present the training dynamics of Qwen3-VL-4B and Qwen3-VL-8B in Figure~\ref{fig:training dynamics}.

\begin{figure}[!h]
    \centering
    \includegraphics[width=\linewidth]{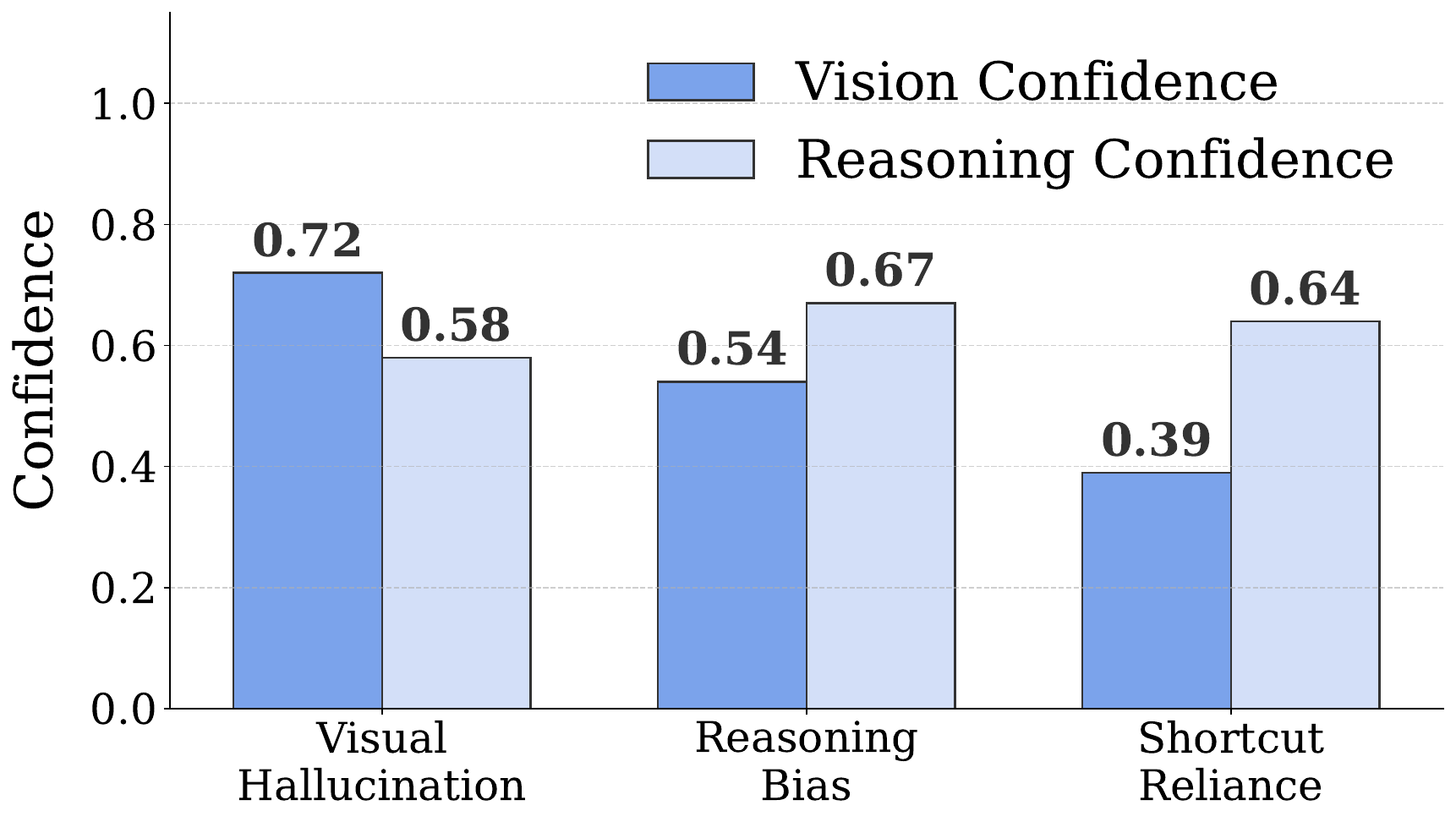}
    \caption{\textbf{Confidence of overconfident wrong answers.} }
    \label{fig:wrong_high}
\end{figure}

\begin{figure}[!h]
    \centering
    \includegraphics[width=\linewidth]{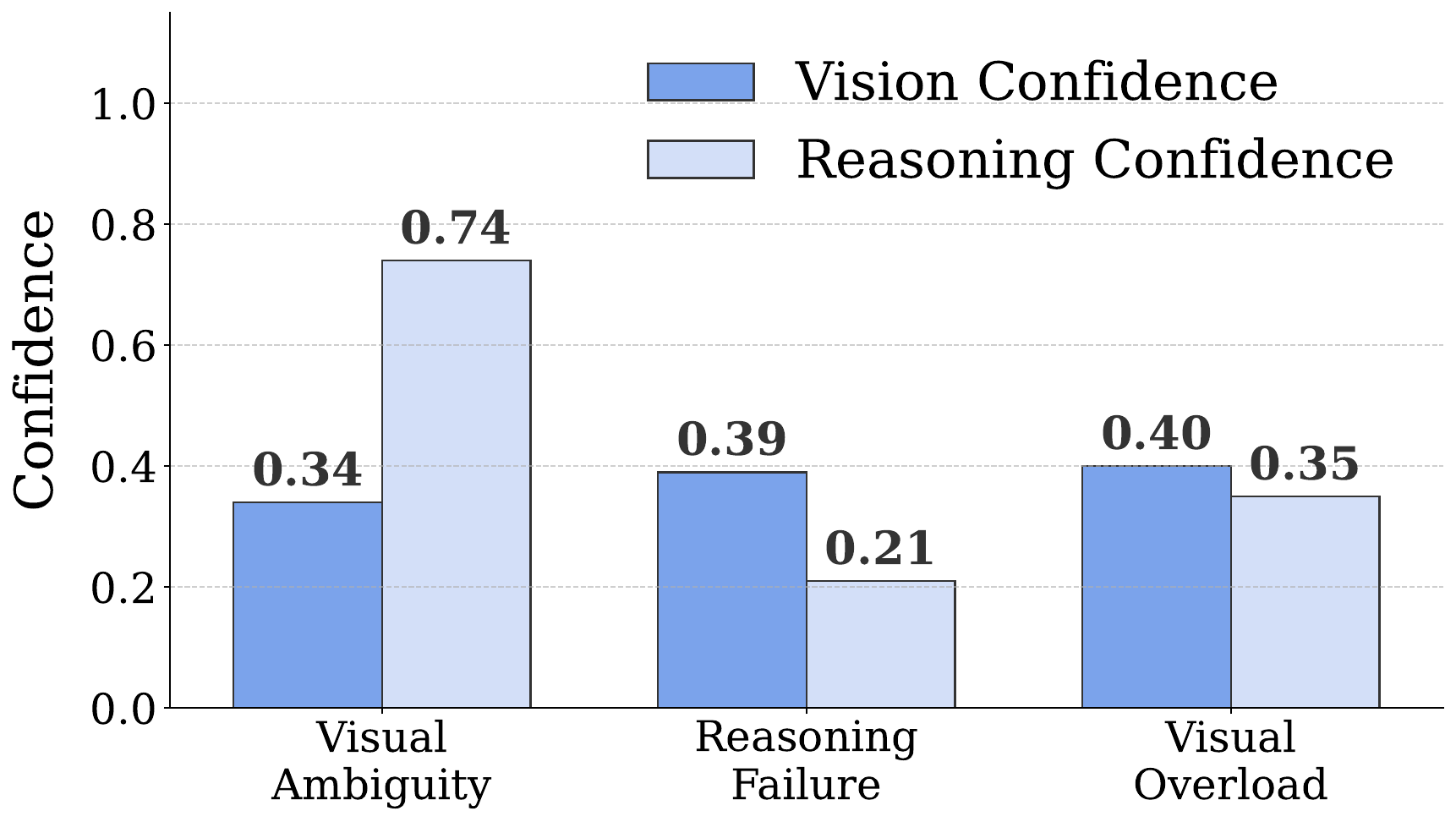}
    \caption{\textbf{Confidence of underconfident correct answers.} }
    \label{fig:error_analysis_correct_2}
\end{figure}

\subsection{Failure Mode Analysis}
\label{appendix: Failure Mode Analysis}

All analyses are conducted on a manually curated subset of 1,000 samples drawn from metric dataset. We systematically reviewed these samples to categorize failure modes based on their holistic confidence scores and prediction correctness.

\textbf{Wrong predictions with high holistic confidence.}
We analyze wrong predictions with high holistic confidence ($> 0.40$), as shown in Figure~\ref{fig:wrong_high}. These cases are categorized into three types. \textit{Visual Hallucination} is the dominant source, where the model confidently infers nonexistent or misperceived visual attributes, leading to incorrect answers and indicating overconfidence. \textit{Reasoning Bias} arises when the model relies on flawed logical shortcuts, producing confident yet incorrect conclusions. The remaining cases involve \textit{Shortcut Reliance}, where prior-driven decision making overrides image-specific evidence, resulting in higher confidence in reasoning rationale.

\textbf{Correct predictions with low holistic confidence.}
We analyze correct predictions with low holistic confidence ($< 0.50$), as shown in Figure~\ref{fig:error_analysis_correct_2}, and categorize them into three types. The majority fall into \textit{Visual Uncertainty}, where critical visual cues are ambiguous or hard to distinguish, leading to lower visual confidence than reasoning confidence. This highlights perception-driven uncertainty captured by our decoupled confidence modeling. The remaining cases include \textit{Reasoning Failure}, where the task surpasses the model's reasoning capacity causing uncertain guesses, and \textit{Visual Overload}, where extremely dense visual inputs make perception difficult. Both latter types reflect increased task difficulty, resulting in uniformly low confidence across visual and reasoning.

\clearpage
\clearpage

\clearpage
\clearpage

\begin{table*}[h]
    \centering
    \caption{\textbf{Comparison of Base Model (Qwen3-VL-4B), Best, and Our Method across various benchmarks.} We report Accuracy, AUROC, and ECE. Best results are highlighted in \textbf{bold}.}
    \label{tab:comparison_grouped}
    \resizebox{\textwidth}{!}{
    \begin{tabular}{lccccccccc}
        \toprule
        \multirow{2}{*}{\textbf{Benchmark}} & \multicolumn{3}{c}{\textbf{Accuracy} $\uparrow$} & \multicolumn{3}{c}{\textbf{AUROC} $\uparrow$} & \multicolumn{3}{c}{\textbf{ECE} $\downarrow$} \\
        \cmidrule(lr){2-4} \cmidrule(lr){5-7} \cmidrule(lr){8-10}
         & Base & Best & Ours & Base & Best & Ours & Base & Best & Ours \\
        \midrule

        \multicolumn{10}{l}{\textit{\textbf{Mathematical and Geometric Reasoning}}} \\
        DynaMath     & 0.486 & 0.718 & \textbf{0.753} & 0.513 & 0.716 & \textbf{0.797} & 0.423 & 0.165 & \textbf{0.081} \\
        Geo3K        & 0.514 & 0.616 & \textbf{0.671} & 0.504 & \textbf{0.801} & 0.792 & 0.773 & 0.159 & \textbf{0.073} \\
        MathVerse    & 0.426 & 0.796 & \textbf{0.807} & 0.416 & 0.659 & \textbf{0.735} & 0.561 & 0.142 & \textbf{0.042} \\
        MathVision   & 0.171 & 0.440 & \textbf{0.483} & 0.501 & \textbf{0.814} & 0.800 & 0.794 & 0.207 & \textbf{0.170} \\
        MathVista    & 0.679 & \textbf{0.772} & 0.730 & 0.566 & 0.710 & \textbf{0.778} & 0.254 & 0.132 & \textbf{0.107} \\
        WeMath       & 0.580 & 0.771 & \textbf{0.820} & 0.593 & 0.647 & \textbf{0.802} & 0.268 & 0.164 & \textbf{0.048} \\
        \midrule

        \multicolumn{10}{l}{\textit{\textbf{Logical Reasoning}}} \\
        LogicVista   & 0.456 & 0.519 & \textbf{0.570} & 0.615 & 0.757 & \textbf{0.794} & 0.315 & 0.232 & \textbf{0.203} \\
        \midrule

        \multicolumn{10}{l}{\textit{\textbf{Visual-Dominant Reasoning}}} \\
        CLEVR        & 0.920 & 0.935 & \textbf{0.935} & 0.517 & 0.577 & \textbf{0.797} & \textbf{0.025} & 0.058 & 0.035 \\
        MathVerse$_V$& 0.283 & 0.748 & \textbf{0.781} & 0.519 & 0.669 & \textbf{0.721} & 0.508 & 0.171 & \textbf{0.056} \\
        \midrule

        \multicolumn{10}{l}{\textit{\textbf{Multi-discipline Reasoning}}} \\
        A-OKVQA      & 0.836 & 0.861 & \textbf{0.875} & 0.584 & 0.592 & \textbf{0.695} & 0.022 & 0.112 & \textbf{0.017} \\
        MMK12        & 0.489 & 0.741 & \textbf{0.747} & 0.468 & 0.651 & \textbf{0.714} & 0.432 & 0.182 & \textbf{0.083} \\
        MMMU-Pro     & 0.249 & 0.436 & \textbf{0.458} & 0.610 & 0.694 & \textbf{0.735} & 0.474 & 0.340 & \textbf{0.335} \\
        ViRL-39K-Test& 0.620 & 0.796 & \textbf{0.816} & 0.406 & 0.729 & \textbf{0.753} & 0.622 & 0.113 & \textbf{0.026} \\

        \midrule
        \textbf{Average} & 0.516 & 0.704 & \textbf{0.727} & 0.524 & 0.694 & \textbf{0.763} & 0.421 & 0.167 & \textbf{0.098} \\
        \bottomrule
    \end{tabular}
    }
\end{table*}

\begin{table*}[ht]
    \centering
    \caption{\textbf{Comparison of Base Model (8B), Best, and Our Method across various benchmarks.} We report Accuracy, AUROC, and ECE. Best results are highlighted in \textbf{bold}.}
    \label{tab:comparison_grouped_8b}
    \resizebox{\textwidth}{!}{
    \begin{tabular}{lccccccccc}
        \toprule
        \multirow{2}{*}{\textbf{Benchmark}} & \multicolumn{3}{c}{\textbf{Accuracy} $\uparrow$} & \multicolumn{3}{c}{\textbf{AUROC} $\uparrow$} & \multicolumn{3}{c}{\textbf{ECE} $\downarrow$} \\
        \cmidrule(lr){2-4} \cmidrule(lr){5-7} \cmidrule(lr){8-10}
         & Base & Best & Ours & Base & Best & Ours & Base & Best & Ours \\
        \midrule

        \multicolumn{10}{l}{\textit{\textbf{Mathematical and Geometric Reasoning}}} \\
        DynaMath     & 0.680 & 0.766 & \textbf{0.784} & 0.576 & 0.667 & \textbf{0.769} & 0.460 & 0.160 & \textbf{0.058} \\
        Geo3K        & 0.514 & 0.621 & \textbf{0.729} & 0.556 & 0.761 & \textbf{0.780} & 0.734 & 0.192 & \textbf{0.056} \\
        MathVerse    & 0.622 & 0.813 & \textbf{0.838} & 0.504 & 0.656 & \textbf{0.742} & 0.372 & 0.129 & \textbf{0.055} \\
        MathVision   & 0.266 & 0.473 & \textbf{0.540} & 0.527 & 0.771 & \textbf{0.815} & 0.428 & 0.249 & \textbf{0.094} \\
        MathVista    & 0.678 & 0.733 & \textbf{0.771} & 0.574 & 0.644 & \textbf{0.753} & 0.459 & 0.198 & \textbf{0.079} \\
        WeMath       & 0.699 & 0.801 & \textbf{0.836} & 0.567 & 0.730 & \textbf{0.777} & 0.388 & 0.110 & \textbf{0.039} \\
        \midrule

        \multicolumn{10}{l}{\textit{\textbf{Logical Reasoning}}} \\
        LogicVista   & 0.508 & 0.600 & \textbf{0.611} & 0.580 & 0.688 & \textbf{0.836} & 0.308 & 0.253 & \textbf{0.109} \\
        \midrule

        \multicolumn{10}{l}{\textit{\textbf{Visual-Dominant Reasoning}}} \\
        CLEVR        & 0.910 & 0.935 & \textbf{0.940} & 0.545 & 0.495 & \textbf{0.723} & 0.332 & 0.069 & \textbf{0.029} \\
        MathVerse$_V$& 0.573 & 0.776 & \textbf{0.804} & 0.502 & 0.660 & \textbf{0.743} & 0.398 & 0.162 & \textbf{0.052} \\
        \midrule

        \multicolumn{10}{l}{\textit{\textbf{Multi-discipline Reasoning}}} \\
        A-OKVQA      & 0.829 & 0.872 & \textbf{0.875} & 0.642 & 0.593 & \textbf{0.691} & 0.057 & 0.107 & \textbf{0.059} \\
        MMK12        & 0.585 & 0.780 & \textbf{0.809} & 0.506 & 0.691 & \textbf{0.777} & 0.301 & 0.131 & \textbf{0.039} \\
        MMMU-Pro     & 0.383 & 0.518 & \textbf{0.522} & 0.579 & 0.634 & \textbf{0.740} & 0.518 & 0.357 & \textbf{0.220} \\
        ViRL-39K-Test& 0.689 & 0.811 & \textbf{0.835} & 0.537 & 0.723 & \textbf{0.783} & 0.460 & 0.109 & \textbf{0.033} \\

        \midrule
        \textbf{Average} & 0.610 & 0.731 & \textbf{0.761} & 0.553 & 0.670 & \textbf{0.764} & 0.401 & 0.171 & \textbf{0.071} \\
        \bottomrule
    \end{tabular}
    }
\end{table*}

\end{document}